\documentclass[11pt]{article}

\usepackage[final]{acl}

\usepackage{times}
\usepackage{latexsym}

\usepackage[T1]{fontenc}

\usepackage[utf8]{inputenc}

\usepackage{microtype}

\usepackage{inconsolata}

\usepackage{graphicx}
\usepackage{microtype}
\usepackage{booktabs}
\usepackage{amsmath}
\usepackage{multirow}
\usepackage{cleveref}
\usepackage{tcolorbox}
\usepackage{array}
\usepackage{alltt}
\usepackage{xcolor}
\usepackage{soul}   
\usepackage{url}
\usepackage{tikz}
\usepackage{fontawesome5}
\usepackage{enumitem}
\usepackage{colortbl}
\usepackage{float}

\usetikzlibrary{arrows.meta, calc, positioning}

\definecolor{phiblue}{HTML}{CFE3F7}
\definecolor{ipisalmon}{HTML}{FBD9CE}
\newcommand{\phis}[1]{%
  \tikz[baseline=(x.base)]{\node[inner xsep=1.5pt, inner ysep=0.5pt, rounded corners=1.5pt,
    fill=cbGen!12, text=cbGenTx] (x) {#1};}}
\newcommand{\ipis}[1]{%
  \tikz[baseline=(x.base)]{\node[inner xsep=1.5pt, inner ysep=0.5pt, rounded corners=1.5pt,
    fill=cbOut!12, text=cbOutTx] (x) {#1};}}
\definecolor{sdgray}{gray}{0.50}
\newcommand{\sd}[1]{\textcolor{sdgray}{\tiny$\pm$#1}}

\definecolor{oiBlue}{HTML}{0072B2}
\definecolor{oiGreen}{HTML}{009E73}
\definecolor{oiOrange}{HTML}{E69F00}
\definecolor{oiPurple}{HTML}{CC79A7} 
\definecolor{cbPatient}{HTML}{0072B2}
\definecolor{cbBoth}{HTML}{009E73}
\definecolor{cbClinician}{HTML}{D55E00}
\definecolor{cbPatientTx}{HTML}{004E7A}
\definecolor{cbBothTx}{HTML}{00614A}
\definecolor{cbClinicianTx}{HTML}{8A3D00}
\definecolor{cbInk}{HTML}{2C2C2A}
\definecolor{cbMuted}{HTML}{6B6B66}
\definecolor{cbGen}{HTML}{0072B2}
\definecolor{cbAnn}{HTML}{009E73}
\definecolor{cbOut}{HTML}{D55E00}
\definecolor{cbChk}{HTML}{7F77DD}
\definecolor{cbGenTx}{HTML}{004E7A}
\definecolor{cbAnnTx}{HTML}{00614A}
\definecolor{cbOutTx}{HTML}{8A3D00}
\definecolor{cbChkTx}{HTML}{4A429E}
\definecolor{cbMuted}{HTML}{6B6B66}

\definecolor{r1}{HTML}{E8F0F7}  
\definecolor{r2}{HTML}{FBE8D3}  
\definecolor{r3}{HTML}{F5CFC0}  
\newcommand{\rone}{\cellcolor{r1}1}
\newcommand{\rtwo}{\cellcolor{r2}2}
\newcommand{\rthree}{\cellcolor{r3}3}

\newcommand{\datasetname}{\textsc{VESTIGE}}

\title{VESTIGE: What De-identification Leaves Behind in Multilingual Healthcare Communication}

\author{
 \textbf{Christoph Otto\textsuperscript{1*}},
 \textbf{Antonin Meyer\textsuperscript{2*}},
 \textbf{Vera Czehmann\textsuperscript{1}},
 \textbf{Matthieu Le Cauchois\textsuperscript{2\textdagger}},
\\
 \textbf{Adèle Lutun\textsuperscript{2\textdagger}},
 \textbf{Katharina Kersting\textsuperscript{1}},
 \textbf{Elisabeth Eberle\textsuperscript{1}},
 \textbf{Sebastian Möller\textsuperscript{1,3}},
\\
 \textbf{Lisa Raithel\textsuperscript{1,3,4*}}
\\
\\
 \textsuperscript{1}Deutsches Forschungszentrum für Künstliche Intelligenz,
 \textsuperscript{2}Doctolib,\\
 \textsuperscript{3}Technische Universität Berlin,
 \textsuperscript{4}Berlin Institute for the Foundations of Learning and Data
\\
 \small{\textsuperscript{*}Equal contribution. \textsuperscript{\textdagger}Work done while at Doctolib.}
\\
 \small{
   \textbf{Correspondence:} \href{mailto:raithel@tu-berlin.de}{raithel@tu-berlin.de}
 }
}

\begin{document}
\maketitle
\begin{abstract}
We present \datasetname, a multilingual benchmark for evaluating automated de-identification and auditing re-identification risk in healthcare-related free text.
The dataset comprises 508 synthetic documents in German, French, and English across eight clinically relevant genres in ambulatory care (e.g., \textit{medical dictation}, \textit{consultation-assistant dialogue}), generated via a two-stage LLM pipeline: a structured blueprint of the document content is sampled, then rendered into natural language under genre- and language-specific style constraints.
All documents are annotated with direct (PII) and indirect (IPI) identifier spans (IAA with $\alpha=0.91$ and $\alpha=0.58$, respectively), and each identifier type is assigned a multi-dimensional re-identification risk profile grounded in GDPR and HIPAA.
We report baseline de-identification performance of F$_1=0.72$ and present a prototype tool that links missed identifiers to their risk profiles to support manual inspection of residual risk. 
The benchmark thus combines three European languages, eight healthcare genres, and a fine-grained, regulation-grounded risk taxonomy for every identifier type. Data, models and code can be found on Hugging Face\footnote{\url{https://hf.co/collections/doctolib-lab/vestige}, \url{https://hf.co/collections/DFKI-SLT/vestige}} and GitHub\footnote{\url{https://github.com/DFKI-NLP/vestige}}.
\end{abstract}

\section{Introduction}\label{sec:intro}

\definecolor{cbPatient}{HTML}{0072B2}
\definecolor{cbBoth}{HTML}{009E73}
\definecolor{cbClinician}{HTML}{D55E00}
\definecolor{cbPatientTx}{HTML}{004E7A}
\definecolor{cbBothTx}{HTML}{00614A}
\definecolor{cbClinicianTx}{HTML}{8A3D00}
\definecolor{cbInk}{HTML}{2C2C2A}
\definecolor{cbMuted}{HTML}{6B6B66}

\begin{figure}[t]
\centering
\begin{tikzpicture}[
  font=\tiny,
  stage/.style={rounded corners=4pt, draw=#1!45, line width=0.4pt,
                fill=cbMuted!6, text=cbInk, minimum width=2.15cm,
                minimum height=0.95cm, align=center, inner sep=3pt},
  doc/.style 2 args={rounded corners=3pt, draw=#1!45, line width=0.4pt,
              fill=#1!8, text=#2, minimum height=0.52cm,
              align=left, text width=2.5cm, inner xsep=5pt, inner ysep=3pt},
  spine/.style={-{Latex[length=1.6mm]}, line width=0.7pt, draw=cbMuted!65},
  link/.style={line width=0.4pt, draw=cbMuted!30},
  swatch/.style={rounded corners=2pt, draw=#1!45, line width=0.4pt,
                 fill=#1!8, minimum width=0.28cm, minimum height=0.28cm,
                 inner sep=0pt}
]

\node[stage=cbPatient]   (s1) at (0,-0.62) {\textbf{Access \& advice}\\[1.5pt]
  {\color{cbMuted}Before or without visit}};
\node[stage=cbBoth]      (s2) at (0,-2.58) {\textbf{Consultation}\\[1.5pt]
  {\color{cbMuted}During visit}};
\node[stage=cbClinician] (s3) at (0,-3.92) {\textbf{Documentation}\\[1.5pt]
  {\color{cbMuted}After visit}};
\node[stage=cbClinician] (s4) at (0,-5.22) {\textbf{Communication}\\[1.5pt]
  {\color{cbMuted}Handover}};
  
\draw[spine] (s1) -- (s2);
\draw[spine] (s2) -- (s3);
\draw[spine] (s3) -- (s4);

\node[doc={cbPatient}{cbPatientTx}]     (d1a) at (3.9, 0.00) {Patient request};
\node[doc={cbPatient}{cbPatientTx}]     (d1b) at (3.9,-0.62) {Phone-assistant transcript};
\node[doc={cbPatient}{cbPatientTx}]     (d1c) at (3.9,-1.24) {AI health assistant message};
\node[doc={cbBoth}{cbBothTx}]           (d2a) at (3.9,-1.96) {Consultation dialogue};
\node[doc={cbBoth}{cbBothTx}]           (d2b) at (3.9,-2.58) {Consultation transcript};
\node[doc={cbBoth}{cbBothTx}]           (d2c) at (3.9,-3.20) {Consultation summary};
\node[doc={cbClinician}{cbClinicianTx}] (d3a) at (3.9,-3.92) {Medical dictation};
\node[doc={cbClinician}{cbClinicianTx}] (d4a) at (3.9,-5.22) {Referral letters};


\draw[link] (s1.east) -- (d1a.west);
\draw[link] (s1.east) -- (d1b.west);
\draw[link] (s1.east) -- (d1c.west);
\draw[link] (s2.east) -- (d2a.west);
\draw[link] (s2.east) -- (d2b.west);
\draw[link] (s2.east) -- (d2c.west);
\draw[link] (s3.east) -- (d3a.west);
\draw[link] (s4.east) -- (d4a.west);

\node[swatch=cbPatient] (l1) at (-0.95,-6.00) {};
\node[anchor=west, text=cbMuted] at (-0.80,-6.00)
  {\includegraphics[height=0.30cm]{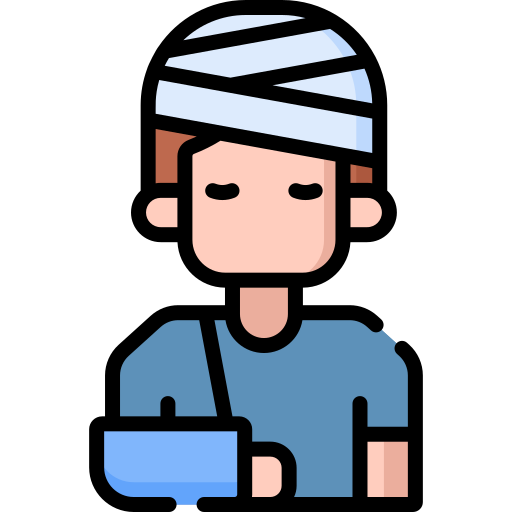}\, patient};
\node[swatch=cbBoth] (l2) at (1.35,-6.00) {};
\node[anchor=west, text=cbMuted] at (1.50,-6.00)
  {\includegraphics[height=0.30cm]{figures/patient2.png}%
   \hspace{-0.5pt}\includegraphics[height=0.30cm]{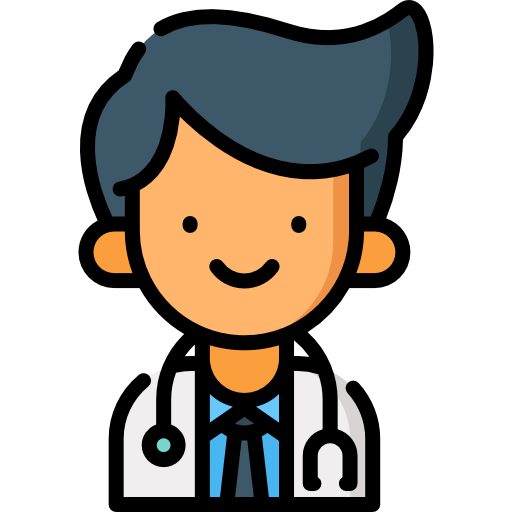}\, both};
\node[swatch=cbClinician] (l3) at (3.55,-6.00) {};
\node[anchor=west, text=cbMuted] at (3.70,-6.00)
  {\includegraphics[height=0.30cm]{figures/doctor.png}\, clinician(s)};

\end{tikzpicture}
\caption{
The eight document genres in VESTIGE along a patient's care path with a general practitioner (or AI), from access and advice before a visit to communication afterwards.
Colors mark whether a genre is produced by the patient (blue), the clinician(s) (orange), or both (green).
}
\label{fig:patient-journey}
\end{figure}

Healthcare runs on written and spoken records of many kinds.
A single patient's care can generate medical dictations, patient requests, phone-assistant transcripts, dialogues and summaries produced by consultation assistants, and referral letters, among others (see \Cref{fig:patient-journey}).
These documents differ in purpose and form, from structured notes to full conversational transcripts, but share one property: they contain personal information about the people involved.
Before such records can be re-used for research, model development, or product evaluation, that information must be removed.

Much of the literature frames this as de-identification \citep{stubbs_annotating_2015}, i.e., the detection and redaction of direct identifiers.
Direct identifiers, or personally identifiable information (PII), identify an individual on their own, e.g., a name or an address, and include the protected health information (PHI) enumerated 
under HIPAA Safe Harbor.\footnote{\url{https://www.hhs.gov/hipaa/}}
Indirect identifiers, or indirect personal information (IPI), identify no one on their own, but can single out a person in combination \citep{baroud_beyond_2025}, e.g., a patient visiting a certain medical facility at a specific time. 
In the European Union the relevant standard is stricter.
Under GDPR Recital 26\footnote{\url{https://tinyurl.com/eu-lex-32016R0679}}, data are anonymous only when the data subject is no longer identifiable by ``means reasonably likely to be used'', 
which does not only refer to direct identifiers but includes IPI as well.

Free-text clinical records carry a great deal of such additional information, for instance a patient's living situation, work, family, and medical history.
These details identify no one on their own, yet in combination they can single out a person \citep{oneill_risks_2016,sweeney_re-identification_2017,janmey_re-identification_2018}.

\definecolor{cbGen}{HTML}{0072B2}
\definecolor{cbAnn}{HTML}{009E73}
\definecolor{cbOut}{HTML}{D55E00}
\definecolor{cbChk}{HTML}{7F77DD}
\definecolor{cbGenTx}{HTML}{004E7A}
\definecolor{cbAnnTx}{HTML}{00614A}
\definecolor{cbOutTx}{HTML}{8A3D00}
\definecolor{cbChkTx}{HTML}{4A429E}
\definecolor{cbMuted}{HTML}{6B6B66}

\begin{figure*}[t]
\centering
\begin{tikzpicture}[
  x=1cm, y=1cm,
  base/.style={rounded corners=3pt, line width=0.4pt},
  art/.style={base, draw=#1!45, fill=#1!8,
              minimum width=2.4cm, minimum height=1.5cm},
  sub/.style={base, draw=cbAnn!45, fill=cbAnn!8,
              minimum width=2.4cm, minimum height=0.6cm},
  badge/.style={base, draw=cbChk!50, fill=cbChk!8,
                minimum width=2.4cm, minimum height=0.85cm},
  wrap/.style={base, draw=cbMuted!45, dash pattern=on 1.6pt off 1.4pt,
               minimum width=2.8cm, minimum height=1.9cm},
  flow/.style={-{Latex[length=1.5mm]}, draw=cbMuted!65, line width=0.65pt},
  loop/.style={-{Latex[length=1.4mm]}, draw=cbChk!75, line width=0.6pt},
  lead/.style={draw=cbMuted!45, line width=0.4pt, dash pattern=on 1.2pt off 1.2pt},
  ttl/.style={font=\scriptsize\bfseries, text=#1, align=center},
  cueN/.style={font=\tiny, text=cbMuted, align=center}
]

\node[art=cbGen] (inputs)    at (1.2,0) {};
\node[art=cbGen] (blueprint) at (4.3,0) {};
\node[art=cbGen] (documents) at (7.4,0) {};
\node[wrap]      (annwrap)   at (10.7,0) {};
\node[sub]       (spans)     at (10.7, 0.35) {};
\node[sub]       (risk)      at (10.7,-0.35) {};
\node[art=cbOut] (bench)     at (13.8,0) {};
\node[badge]     (valb)      at (4.3,-1.775) {};
\node[badge]     (qualb)     at (7.4,-1.775) {};

\draw[flow] (2.4,0) -- (3.1,0);
\draw[flow] (5.5,0) -- (6.2,0);
\draw[flow] (8.6,0) -- (9.3,0);
\draw[flow] (12.1,0) -- (12.6,0);

\draw[lead] (4.3,-0.75) -- (4.3,-1.35);
\draw[lead] (7.4,-0.75) -- (7.4,-1.35);

\draw[loop] (3.1,-1.775) .. controls (2.65,-1.775) and (2.65,-0.55) .. (3.1,-0.55);
\draw[loop] (6.2,-1.775) .. controls (5.75,-1.775) and (5.75,-0.55) .. (6.2,-0.55);

\node[ttl=cbGenTx] at (1.2, 0.40) {Inputs};
\node[cueN]        at (1.2, 0.02) {templates, anchors};
\node[cueN]        at (1.2,-0.30) {and priors};

\node[ttl=cbGenTx] at (4.3, 0.40) {Blueprint};
\node[cueN, anchor=west] at (3.78, 0.02) {LLM call 1};
\node[cueN]        at (4.3,-0.30) {structured JSON};

\node[ttl=cbGenTx] at (7.4, 0.40) {Documents};
\node[cueN, anchor=west] at (6.88, 0.02) {LLM call 2};
\node[cueN]        at (7.4,-0.30) {text rendering};

\node[cueN] at (10.7, 1.12) {Annotation};
\node[font=\tiny, text=cbAnnTx] at (10.7, 0.35) {Token spans};
\node[font=\tiny, text=cbAnnTx] at (10.7,-0.35) {Risk profiles};

\node[ttl=cbOutTx] at (13.8, 0.15) {Benchmark};
\node[cueN]        at (13.8,-0.15) {508 documents};

\node[font=\tiny, text=cbChkTx, anchor=west] at (3.78,-1.65) {Validation};
\node[cueN] at (4.3,-1.95) {schema + retry};
\node[font=\tiny, text=cbChkTx, anchor=west] at (6.78,-1.65) {Quality check};
\node[cueN] at (7.4,-1.95) {plausibility + fidelity};

\begin{scope}[shift={(3.65,0.02)}, line join=round]
  \draw[cbGen, line width=.5pt]
    (0,.13) .. controls (.03,.04) and (.09,.02) .. (.16,0)
            .. controls (.09,-.02) and (.03,-.04) .. (0,-.13)
            .. controls (-.03,-.04) and (-.09,-.02) .. (-.16,0)
            .. controls (-.09,.02) and (-.03,.04) .. cycle;
\end{scope}
\begin{scope}[shift={(6.75,0.02)}, line join=round]
  \draw[cbGen, line width=.5pt]
    (0,.13) .. controls (.03,.04) and (.09,.02) .. (.16,0)
            .. controls (.09,-.02) and (.03,-.04) .. (0,-.13)
            .. controls (-.03,-.04) and (-.09,-.02) .. (-.16,0)
            .. controls (-.09,.02) and (-.03,.04) .. cycle;
\end{scope}

\begin{scope}[shift={(10.00,0.35)}, line cap=round]
  \draw[cbAnn, line width=.45pt] (-.11,0) -- (.11,0);
  \draw[cbAnn, line width=.55pt]
      (-.03,.05)--(-.03,-.05)
      (-.03,.05)--(-.01,.05)
      (-.03,-.05)--(-.01,-.05);
  \draw[cbAnn, line width=.55pt]
      (.03,.05)--(.03,-.05)
      (.03,.05)--(.01,.05)
      (.03,-.05)--(.01,-.05);
\end{scope}

\begin{scope}[shift={(10.00,-0.35)}, line cap=round]
  \draw[cbAnn, line width=.45pt] (-.10,-.08) -- (-.10,.00);
  \draw[cbAnn, line width=.45pt] (-.03,-.08) -- (-.03,.08);
  \draw[cbAnn, line width=.45pt] (.04,-.08) -- (.04,.04);
  \draw[cbAnn, line width=.45pt] (-.13,-.08) -- (.08,-.08);
\end{scope}

\begin{scope}[shift={(3.62,-1.65)}, line join=round, line cap=round]
  \draw[cbChk, line width=.5pt]
    (0,.15) -- (.11,.10) -- (.11,-.02) -- (0,-.15) -- (-.11,-.02) -- (-.11,.10) -- cycle;
  \draw[cbChk, line width=.7pt] (-.055,.01) -- (-.012,-.035) -- (.07,.06);
\end{scope}
\begin{scope}[shift={(6.72,-1.65)}, line join=round, line cap=round]
  \draw[cbChk, line width=.5pt]
    (0,.15) -- (.11,.10) -- (.11,-.02) -- (0,-.15) -- (-.11,-.02) -- (-.11,.10) -- cycle;
  \draw[cbChk, line width=.7pt] (-.055,.01) -- (-.012,-.035) -- (.07,.06);
\end{scope}

\end{tikzpicture}
\caption{Dataset construction. 
Templates, anchors and priors (\textit{Inputs}) are turned into a JSON \textit{Blueprint} (LLM call 1), checked against the genre schema (\textit{Validation}), rendered into text (\textit{Documents}, LLM call 2), checked for plausibility and fidelity (\textit{Quality check}), and annotated with spans and risk profiles (\textit{Annotation}).
}
\label{fig:dataset-construction}
\end{figure*}
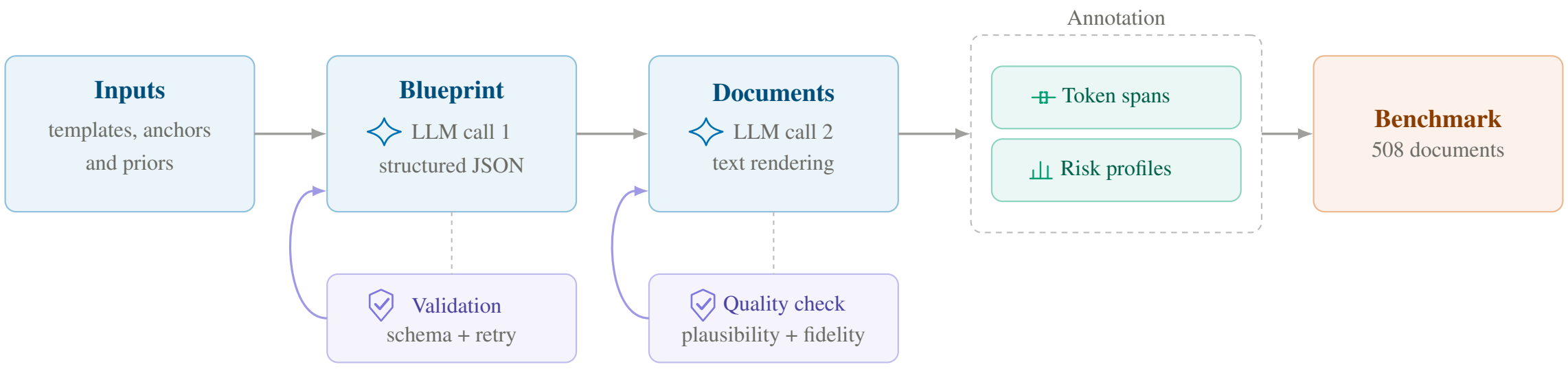%

Benchmarks are what let systems be compared and hold industry and academia to the same standard, but the coverage of existing ones is uneven.
English resources dominate \citep{stubbs_annotating_2015,johnson_mimic-iii_2016,johnson_mimic-iv_2023}, and the corpora in other languages described in the literature are individually monolingual \citep{marimon_automatic_2019,lohr_-identifying_2024,grouin_-identification_2014}.
Where indirect identifiers are annotated at all, this is done either outside the clinical domain \citep{pilan_text_2022} or on translated clinic-authored documents \citep{baroud_multigrascco_2026}, leaving patient-facing genres uncovered.
Annotation schemes are mostly informed by data protection regulation \citep{marimon_automatic_2019}, but they specify which categories must be masked rather than how much re-identification risk each category carries.
To our knowledge, no clinical de-identification benchmark releases per-identifier-type risk scores alongside its annotations.
We address this with \datasetname, a fully synthetic, human-annotated benchmark comprising:
\begin{description}[leftmargin=*]
  \item[Multilingual, multi-genre coverage:] Eight healthcare genres in German, French, and English, spoken (then transcribed) and written, as used in production by a European digital-health provider.
  \item[Two-level identifier taxonomy:] Span-level annotation of direct and indirect identifiers.
  \item[Regulation-grounded risk scores:] Each identifier type is scored on five dimensions (1--3), yielding a composite re-identification risk score.
  \item[Transparent generation pipeline:] A re-runnable two-stage pipeline from a structured blueprint, i.e., a JSON outline of the intended document content, to LLM rendering, based on country-specific demographics.
  \item[Residual-risk inspection tool:] An interactive demonstrator that links missed identifiers to the risk profile of their type and flags documents for manual review.
\end{description}

\section{Benchmark Construction}

The following describes the construction of the synthetic dataset, based on production data provided by a European healthcare provider. 
We further base the synthetization on country-specific statistics and direct and indirect identifiers from related works.

\subsection{Source Data Analysis}
\label{sec:source_data_analysis}

The corpus covers eight genres (\Cref{fig:patient-journey}).
\emph{Medical dictation} contains clinical dictation transcripts in note style.
\emph{Patient requests} are short patient messages asking for appointments, and \emph{phone assistant transcripts} record patient calls handled by a phone assistant.
Three genres come from consultation assistants: \emph{consultation assistant dialogue} in dialogue form, \emph{consultation assistant summary} as a brief structured overview, and \emph{consultation assistant transcript} as a verbatim record preserving turns and disfluencies.
\emph{Referral letters} are addressed to other healthcare professionals, and \emph{AI health assistant} messages are short caregiver questions about child health and development.

To guide generation, we analyzed 150 seed documents in German and French (53 German and 97 French).
\textit{Consultation assistant} documents, and \textit{patient requests} were produced by in-house employees acting as patients or synthesized by the provider for production testing, and therefore contain no real patient data. 
The remaining documents were anonymized and manually reviewed on-premise by the healthcare provider before transfer: automatic anonymization first, then independent review by two data scientists for remaining PII and IPI, and a final inspection by a third reviewer. 
No original document was sent to any third-party service. 
The style anchors used during generation (\Cref{sec:pipeline}) were written by three of the authors, a researcher with a PhD in clinical NLP, a master's level computational linguist, and a data scientist from the healthcare provider.
Each anchor is based on reviewed seed documents of one genre, reusing some formulations while changing the clinical content, e.g., medications or symptoms.

We first analyzed the source documents with respect to their communicative register, average length and vocabulary to derive genre-specific generation prompts and style anchors. The de-identified placeholders in the reviewed seed documents (e.g., \textsc{$<$Name$>$)} were replaced with values using the \texttt{Faker} library.\footnote{\url{https://github.com/joke2k/faker}} 
We further labeled them automatically with models provided by related work \citep{baroud_multigrascco_2026}, to see which identifiers occur in the data.
Note that these models used slightly different label sets for training, so we take the results as indicative only.
Among PII, date and person mentions make up about 85\% of all detected spans, and among IPI, time expressions and references to healthcare facilities make up a similar share (on the seed data, with the label set of \citet{baroud_multigrascco_2026}).

\subsection{Synthetic Data Generation Pipeline}
\label{sec:pipeline}

\Cref{fig:dataset-construction} summarizes the pipeline for the construction of the data set. 
Corpus generation proceeds in two LLM calls: the first constructs a blueprint, i.e., a structured JSON representation of the intended document, similar to the clinical vignettes, i.e., short descriptions of a patient case, that \citet{nastasi_vignette-based_2023} used to evaluate medical advice by ChatGPT,
and the second renders this representation into genre- and language-specific clinical text.
The resulting corpus is subsequently annotated with PII and IPI spans and augmented with type-level risk profiles, as described in \Cref{sec:risk}. 
All LLM calls use \texttt{GPT-5.3-chat} \cite{openai2025gpt5card}.
To make the pipeline transparent and re-runnable, the complete prompts and hyperparameters of both stages are given in \Cref{app:prompt_template}.\footnote{Note that API outputs are not guaranteed to be identical across runs, even at temperature 0.}

\paragraph{Generation inputs and blueprint construction (\Cref{fig:dataset-construction}, \textit{Inputs}, \textit{Blueprint}).}
For each document, we sampled four profile attributes (age group, gender, smoking status, and employment status) from publicly available country-specific prior distributions\footnote{All used databases are linked in the accompanying repository.} (Germany, France, UK). 
The LLM received the target genre and language, the sampled profile attributes, a genre-specific document template listing the required and optional fields, the target-length range and one genre anchor, returning a structured JSON blueprint.
The blueprint contained values for the genre specific fields as well as a shared clinical outline describing the visit context, chief complaint, relevant clinical information, clinical impression and care plan.

As part of the generation pipeline, each blueprint was validated automatically (\textit{Validation}). 
All required fields had to be non-empty and each blueprint could contain only fields defined in the corresponding document template. Blueprints that failed these checks were regenerated up to three times; documents that did not pass validation after the final attempt were excluded.

We used one reconstructed style anchor per document genre and language. 
These anchors were derived from anonymized seed documents.  

\paragraph{Blueprint-guided text realization (\textit{Documents}).}
A second LLM call rendered each validated blueprint as a clinical text. The prompt specified the target language and genre, the corresponding token-length ranges and the semantic content of the blueprint. The genre anchor used during blueprint construction was supplied again as a reference for structure, register and formatting. 
French documents used French anchors for all genres. 
German documents used German anchors where available and French anchors for the remaining genres. 
English documents used both French and German anchors. When the anchor language differed from the target language, documents were generated cross-lingually using the anchor as a reference.

\paragraph{Quality control and entity diversification (\textit{Quality check}).}
Corpus-level inspection additionally revealed repetitive physician and healthcare-facility names across documents. We therefore identified physician and facility entities that occur in at least two documents and assigned them replacements from curated country-specific name pools.
This procedure replaced 269 physician-name mentions in 59 documents and 283 facility mentions in 159 documents.

\paragraph{Dataset statistics.}
Generation began with a minimum of seven synthetic documents per genre. Additional documents were sampled to approximate the genre distribution of the source data. The final corpus contains 508 documents (170 DE, 170 EN, 168 FR), distributed across eight genres.
For each language, the corpus includes 54 \textit{patient requests}, 26 \textit{medical dictations}, 20 \textit{AI health assistant (AIHA) requests}, 18 \textit{consultation-assistant dialogues}, 16 \textit{consultation-assistant summaries}, 16 \textit{phone-assistant transcripts}, and 10 \textit{consultation-assistant transcripts} and \textit{referral letters}. As shown in \Cref{tab:average-sent-tok-by-genre}, document length varies by genre, from short \textit{patient requests} to verbose \textit{consultation-assistant transcripts}.
Two French documents, one \textit{consultation-assistant transcript} and one \textit{referral letter}, did not pass the final quality checks and were discarded.
\Cref{fig:output_example} shows an example of a generated document (shortened). 
Note that identifiers frequently refer to several people at once, as in \Cref{fig:output_example}, so residual risk has to be assessed at the document level rather than per person.

\begin{figure}[t]
\centering
\begin{tcolorbox}[
    width=0.97\linewidth,
    colback=cbMuted!4,
    colframe=cbMuted!45,
    coltitle=cbMuted!20!black,
    colbacktitle=cbMuted!10,
    boxrule=0.4pt,
    arc=3pt,
    left=5pt, right=5pt, top=4pt, bottom=4pt,
    fonttitle=\scriptsize\bfseries,
    title=Example (AI health assistant)
]
\small
Hello, I'm \phis{72} and still \ipis{working part-time}. I look after my
\phis{4-month-old} \ipis{granddaughter} in \phis{Leeds} when \ipis{my daughter}
is \ipis{at work}. \ipis{Some afternoons} she cries for quite a long time
even though she's been fed and changed. Is this normal at her age, and
are there safe ways I can help calm her when it happens?
\end{tcolorbox}
\caption{Synthetic request to an AIHA from our dataset, with direct (blue) and indirect (orange) identifiers.}
\label{fig:output_example}
\end{figure}

\begin{table}[t]
  \centering
  \small
  \begin{tabular}{lrrr}
    \toprule
    \textbf{Genre} & \textbf{Docs} & \textbf{Avg.\ Sent.} & \textbf{Avg.\ Tok.} \\
    \midrule
    AIHA                 & 60  & 4.7   & 62.3  \\
    Consult.-asst.\ dial.& 54  & 44.0  & 435.2 \\
    Consult.-asst.\ sum. & 48  & 22.0  & 306.3 \\
    Consult.-asst.\ trans.& 29 & 107.2 & 939.8 \\
    Referral letters     & 29  & 19.6  & 296.6 \\
    Medical dictation    & 78  & 36.0  & 548.5 \\
    Patient request      & 162 & 4.3   & 47.7  \\
    Phone assistant      & 48  & 52.0  & 510.3 \\
    \midrule
    All genres & 508 & 26.4 & 300.8 \\
    \bottomrule
  \end{tabular}
  \caption{Average sentences and tokens by genre across all languages.}
  \label{tab:average-sent-tok-by-genre}
\end{table}

\subsection{Assessment of Data Quality}
\label{sec:analysis}

We assess generation quality through (i) genre fidelity in terms of length and structure, (ii) semantic similarity to style anchors, and (iii) a human evaluation with respect to plausibility and audience-specific utility.

\subsubsection{Qualitative Assessment}
 
The generated documents appear generally fluent and semantically coherent, with genre-specific conventions broadly retained. 
In particular, dialog-oriented genres, such as \textit{phone assistant} and \textit{consultation transcript}, exhibit consistent speaker turns. 
Nevertheless, the assessment also reveals two recurring deviations: Length mismatches and structural hallucination.
 
\paragraph{Length mismatch.}
Several genres, including \textit{consultation summary}, \textit{phone assistant}, and \textit{AIHA}, produce longer documents than their seed references. 
Patient requests are longer in German (word ratio 1.54) but slightly shorter in French (0.83), and consultation-assistant dialogues are considerably shorter in both languages (0.51 and 0.26).
\Cref{tab:lengths} (App.) reports word-count ratios between generated and seed documents and the mean token length of the synthetic documents.
For example, \textit{consultation summary} length increases on average from 133 to 254 tokens in German and from 167 to 302 tokens in French, likely partly because demographic and socioeconomic values from the blueprint are verbalized in the generated text.
 
\paragraph{Structural hallucination.} Turn-taking markers occasionally appear in genres where they are absent in the reference data (e.g., \textit{consultation assistant}), suggesting that the model overgeneralizes dialogue conventions. 
Although these markers do not necessarily alter coherence, they reduce the fidelity to the intended genre structure.
 
\paragraph{Semantic similarity.}
We compare each document to its style anchor and to randomly selected anchors, using cosine similarity over multilingual embeddings\footnote{\url{https://hf.co/jinaai/jina-embeddings-v3}}; full results and complementary BERTScore are in \Cref{app:details_generated_docs}.
Synthetic documents are closer to their own anchor than to random ones in 22 of 24 language--genre combinations (but do not copy them), for instance 0.713 versus 0.335 for English \textit{consultation summaries} and 0.759 versus 0.410 for French \textit{phone-assistant} documents.
Both exceptions are \textit{AIHA} documents (DE 0.171 vs.\ 0.243, FR 0.104 vs.\ 0.224), where short seed questions (14.4 tokens on average) are expanded into much longer texts (63 tokens).
Therefore, overall, the generated documents follow their anchors stylistically.

\subsubsection{Human Evaluation}

We complement the automatic evaluation with a human pilot study involving the two audiences of the generated documents, i.e., patients and clinicians. Since these groups interact with clinical text for different purposes, we designed separate audience-specific questionnaires. The patient-facing questionnaire focuses on understandability (P1), actionability (P2), and trustworthiness (P3--P5), while the clinician-facing questionnaire assesses clinical realism (C1), internal consistency (C2), and clinical utility (C3--C5). 
The questionnaire items were informed by established health-information evaluation frameworks, adapting concepts from DISCERN \cite{Charnock105}, PEMAT \cite{SHOEMAKER2014395} and ELF-Q \cite{clerehan2016elfq} to the evaluation of synthetic clinical documents.

Each participant received instructions describing the intended document context and the meaning of the response scale. All items were rated on a five-point Likert scale, with an additional ``Cannot say'' option for dimensions that were not applicable to a particular document. This accommodates the diversity of document genres, as not every criterion applies equally across all document types. Across all assessments, 190 of 750 responses were ``Cannot say''. These responses were excluded from the calculation of mean ratings.

Three participants completed the patient questionnaire and two completed the clinician questionnaire, each rating 30 randomly selected documents (\Cref{tab:userstudy}). Although this pilot allows only a preliminary assessment, there is a tendency towards strong authenticity of the generated documents.  
However, we also received feedback (both from patients and clinicians) that the generated documents were almost ``too perfect'' and that some topics (e.g., tobacco use) were discussed very often. The complete wording of all questionnaire items is provided in \Cref{app:user_study_details}. 

\begin{table}[t]
\centering
\small
\begin{tabular}{p{5.6cm}r}
\toprule
\textbf{Patient-facing evaluation} & \textbf{Mean} \\
\midrule
P1 Understanding what is happening and why & 4.23 \\
P2 Knowing what to do next & 4.42 \\
P3 Feeling concerns were understood & 4.41 \\
P4 Trustworthiness & 4.59 \\
P5 Overall helpfulness & 3.95 \\
\midrule
\textbf{Clinician-facing evaluation} & \textbf{Mean} \\
\midrule
C1 Clinical realism & 4.70 \\
C2 Internal consistency & 4.71 \\
C3 Supports appropriate next clinical step & 4.41 \\
C4 Completeness for follow-up & 4.20 \\
C5 Overall suitability for clinical practice & 4.42 \\
\bottomrule
\end{tabular}
\caption{Average ratings on a 5-point Likert scale. Higher values indicate more positive evaluations.}
\label{tab:userstudy}
\end{table}

\section{Annotation of Identifiers \& Risk}
\label{sec:annotation}

We carried out two annotation tasks at different levels of granularity. 
The first is a \textit{span-level} annotation of direct and indirect identifiers within the generated documents. 
The second assigns a re-identification risk score to each identifier \textit{type}. 

\subsection{Annotation of Identifiers}
\label{sec:annotation_ids}

We distinguish direct (PII) and indirect (IPI) identifiers as defined in \Cref{sec:intro} (\Cref{tab:identifiers-short}).
Our PII taxonomy includes 11 types and our IPI taxonomy 17 types.
Of these, 7 PII and 8 IPI types occur in the final corpus (\Cref{tab:annot_stats}).
The five health-related types (\textsc{Disease} - \textsc{History}) are restricted to rare or unusual clinical content (\Cref{tab:identifiers}), since common clinical content does not single out a patient, and the generated documents contain none.

\begin{table}[h]
\centering
\small
\setlength{\tabcolsep}{6pt}
\renewcommand{\arraystretch}{1.1}
\begin{tabular}{@{}ll@{}}
\toprule
\textbf{PII} & \textbf{IPI} \\
\midrule
\textsc{Name}          & \textsc{Appearance} \\
\textsc{Date}          & \textsc{Time} \\
\textsc{Age}           & \textsc{Age} \\
\textsc{Geolocation}   & \textsc{Geolocation} \\
\textsc{Facility}      & \textsc{Facility} \\
\textsc{Organization}  & \textsc{Disease\_Diagnosis\_Proced.} \\
\textsc{Contact\_Phone}  & \textsc{Symptoms} \\
\textsc{Contact\_Online} & \textsc{Medication} \\
\textsc{Code}          & \textsc{Result} \\
\textsc{Biometric}     & \textsc{History} \\
\textsc{Profession}    & \textsc{Legal} \\
                       & \textsc{Lifestyle} \\
                       & \textsc{Context} \\
                       & \textsc{Demographics} \\
                       & \textsc{Dem.\_Family\_Household} \\
                       & \textsc{Dem.\_Socio} \\
                       & \textsc{Quantity} \\
\bottomrule
\end{tabular}
\caption{Identifier taxonomy (11 PII, 17 IPI) of which 7 PII and 8 IPI occur in the corpus (\Cref{tab:annot_stats}). 
Full definitions and examples in \Cref{tab:identifiers}. 
Some identifiers, such as \textsc{age} occur in both categories, since they can be referred to directly and indirectly. 
}
\label{tab:identifiers-short}
\end{table}

We based the annotation guidelines on related work \citep{pilan_text_2022,lohr_-identifying_2024,baroud_beyond_2025} and on an initial analysis of the original anonymized data described above.\footnote{Annotation guidelines are published on GitHub.} 
Before the main annotation, annotators were trained on the guidelines, which were refined iteratively until the annotators reached satisfying agreement on their application to the training examples.

In total, four human annotators contributed to the task.
Their backgrounds with respect to education, language, and compensation are described in \Cref{app:anno_backgrounds}.
Each document was independently annotated by at least two annotators using the open-source software INCEpTION \citep{klie_inception_2018}. 
Curation was then carried out by the two most experienced annotators, working independently over complementary language pairs (French and German; German and English). 
Inter-annotator agreement (IAA) was assessed separately for PII and IPI using unitizing Krippendorff's $\alpha$\footnote{Calculated with DKPro Agreement 2.2.1.} \citep{krippendorff_bivariate_1970} and span-level $F_1$ \citep{hripcsak_agreement_2005}.
We report both macro- and micro-averaged $F_1$ under both exact span matching and relaxed matching, in which two spans count as a full match if they overlap by at least one token.

\subsubsection{Inter-Annotator Agreement (Identifiers)}

\Cref{tab:iaa_overall} (App.) presents the overall results.
Annotators reached an agreement of $\alpha_{\text{unitizing}}=0.907$ (macro $F_{1, relaxed}=0.927$) and $0.584$ (macro $F_{1, relaxed}=0.669$) for PII and IPI, respectively.
Detailed breakdowns by document genre, language, and individual identifier type are provided in \Crefrange{tab:iaa_overall}{tab:iaa_identifier}.

Across all metrics, agreement was consistently higher for PII than for IPI.
Relaxed matching increased agreement particularly strongly for IPI, indicating that annotators frequently identified the same general textual region but selected different span boundaries.
Nevertheless, relaxed IPI agreement remained below the corresponding PII scores.
Boundary variation therefore does not fully account for this gap, the remaining disagreement is also consistent with additional conceptual ambiguity in deciding  whether contextual information qualifies as potentially identifying.

These patterns were broadly stable across languages and document genres.
PII agreement showed limited variation between languages.
At the genre level, PII agreement was particularly high for \textit{consultation summaries}, while relaxed matching substantially increased agreement in several genres.
IPI annotation remained more difficult across all genres.

\subsubsection{Corpus-Level Annotation Statistics}
 
\Cref{tab:annot_stats} provides identifier frequency statistics in the final annotated corpus. Overall, the corpus contains 3,542 PII spans and 5,922 IPI spans across the three languages. Among PII categories, \textsc{Name} is the most frequent, while \textsc{Time} is the dominant IPI category, reflecting the frequent use of temporal expressions in clinical documents, for example, to describe symptom onset (e.g., \textit{``since yesterday evening''}). 
Of the three \textsc{Demographic} types, only the \textsc{Family\_Household} and \textsc{Socioeconomic} subtypes occur in the corpus, since no document mentions, e.g., ethnicity or sexual orientation.
The overall distributions are comparable across German, English and French, supporting balanced multilingual training and evaluation.

\begin{table}[t]
  \centering
  \small
  \begin{tabular}{lrrrr}
    \toprule
    \textbf{Type} & \textbf{DE} & \textbf{EN} & \textbf{FR} & \textbf{Total} \\
    \midrule
    \multicolumn{5}{l}{\textit{PII spans}} \\
    \textsc{Name}              & 650  & 646  & 467  & 1763 \\
    \textsc{Date}              & 168  & 173  & 167  & 508  \\
    \textsc{Age}               & 38   & 26   & 26   & 90   \\
    \textsc{Geolocation}       & 208  & 194  & 212  & 614  \\
    \textsc{Facility}          & 135  & 156  & 144  & 435  \\
    \textsc{Contact\_Phone}    & 12   & 16   & 17   & 45   \\
    \textsc{Profession}        & 28   & 34   & 25   & 87   \\
    \midrule
    Total PII         & 1239 & 1245 & 1058 & 3542 \\

    \midrule
    \multicolumn{5}{l}{\textit{IPI spans}} \\
    \textsc{Time}                       & 1483 & 1045 & 722  & 3250 \\
    \textsc{Age}                        & 34   & 105  & 82   & 221  \\
    \textsc{Geolocation}                & 0    & 4    & 1    & 5    \\
    \textsc{Lifestyle}                  & 298  & 353  & 234  & 885  \\
    \textsc{Context}                    & 1    & 2    & 6    & 9    \\
    \textsc{Dem.\_Family\_Hou.}          & 57   & 222  & 94   & 373  \\
    \textsc{Dem.\_Socio}                      & 338  & 452  & 279  & 1069 \\
    \textsc{Quantity}                   & 0    & 26   & 84   & 110  \\
    \midrule
    Total IPI                           & 2211 & 2209 & 1502 & 5922 \\
    \bottomrule
  \end{tabular}
  \caption{Total PII/IPI spans in the final annotated corpus. Identifiers from the taxonomy that do not occur in the corpus are not shown.}
  \label{tab:annot_stats}
\end{table}

\subsection{Annotation of Risk}
\label{sec:risk}
We assign risk scores to identifier types to measure the residual risk after automatic detection of PII and IPI in a document.
Every mention a system fails to detect is assigned its pre-defined risk and flagged for manual inspection.

\subsubsection{Risk Dimensions}
Following \citet{prasser_importance_2016} and aligned with GDPR and HIPAA, two annotators scored each identifier type on five dimensions (1 = low, 3 = high risk):
\begin{enumerate}[itemsep=1pt, parsep=0pt, topsep=2pt, leftmargin=*]
  \item \textbf{Membership}: Does the identifier reveal that the person is part of the dataset or a particular care setting?
  \item \textbf{Uniqueness}: How likely is the identifier to single out the patient at a realistic population scale?
  \item \textbf{Stability}: How persistent is the identifier over time and across systems?
  \item \textbf{Linkability}: Can the identifier be joined with external public sources?
  \item \textbf{Legal sensitivity}: Where is the identifier positioned relative to HIPAA/GDPR?
\end{enumerate}
 
\noindent The composite score is the maximum across the five dimensions, so one high-risk dimension places an identifier at the highest level.
PII types receive a minimum legal-sensitivity score of 3 by definition and therefore an aggregated score of 3 throughout.
Documents still containing one direct or three or more indirect identifiers after automatic detection are flagged for expert review, following \citet{hrynaszkiewicz_preparing_2010}.

\subsubsection{Annotation Quality (Risk)}

We report inter-annotator agreement using Krippendorff's $\alpha_{risk}$ (ordinal) as the primary metric and linearly weighted Cohen's $\kappa_{risk}$ as a secondary measure (\Cref{tab:iaa_risk_detailed}, \Cref{app:risk_profiles}).
Agreement was highest for membership ($\alpha_{risk}=0.852$) and linkability (0.742), moderate for uniqueness (0.667) and stability (0.609), and lowest for legal sensitivity (0.579), with IPI using the scale more consistently than PII.
Notably, PII membership shows a near-zero $\alpha_{risk}$ ($-$0.050) despite 81.8\% exact agreement, a prevalence-bias artifact of chance-corrected metrics under skewed distributions \citep{feinstein_high_1990,cicchetti_high_1990}, since PII by definition discloses cohort membership.
A related pooling effect explains why overall values for membership and linkability exceed both subgroup values (\Cref{app:risk_profiles}).
Curation depended on the size of the disagreement (\Cref{tab:curated_risk}): of 140 decisions, 71.4\% were exact matches, 27.9\% one-step disagreements resolved to the higher, more privacy-protective score, and one two-step disagreement was expert-adjudicated.

\subsubsection{Resulting Risk Profiles}
\Cref{tab:risk_example} shows three examples, with the full profiles in \Cref{tab:gold_risk_assessment} (\Cref{app:risk_profiles}).
An indirect location (\textit{``biggest hospital in ...''}) scores low on \textit{uniqueness}, since a vague reference rarely singles anyone out, but high on \textit{stability} and \textit{legal sensitivity}, and so still reaches 3, whereas an indirect time expression stays at 2 across all dimensions.
\textsc{Name} scores 2 on \textit{uniqueness} despite being directly identifying, since common names are shared by many people, which is precisely the kind of divergence between dimensions the scheme is meant to capture.

\begin{table}[h]
\centering
\small
\setlength{\tabcolsep}{4pt}
\begin{tabular}{@{}lcccccc@{}}
\toprule
\textbf{Identifier} & \textbf{Mem} & \textbf{Uni} & \textbf{Sta} & \textbf{Lnk} & \textbf{Leg} & \textbf{Agg} \\
\midrule
\textsc{Name} (PII)    & \rthree & \rtwo  & \rthree & \rthree & \rthree & \rthree \\
\textsc{Time} (IPI)    & \rone   & \rtwo  & \rtwo   & \rtwo   & \rone   & \rtwo   \\
\textsc{Geoloc.} (IPI) & \rtwo   & \rone  & \rthree & \rtwo   & \rthree & \rthree \\
\bottomrule
\end{tabular}
\caption{Example risk profiles for three identifier types, direct mentions of names and indirect mentions of time expressions and geolocations. Cells are shaded by risk level (1 = low, 3 = high). Mem = membership, Uni = uniqueness, Sta = stability, Lnk = linkability, Leg = legal sensitivity, Agg = aggregated risk score.}
\label{tab:risk_example}
\end{table}

\section{Baseline Experiments}
\label{sec:baselines}
We evaluate two multilingual encoders \texttt{XLM-RoBERTa-large} \cite{xlmroberta2019} and
\texttt{Multilingual ModernBERT-base} \cite{marone2025mmbertmodernmultilingualencoder}, which differ in size (large vs.\ base), so their comparison is not size-matched. We trained those models on German, English, French, and multilingual training data, and evaluated each on the corresponding test set.
As a reference point, we also report zero-shot \texttt{GPT-5.4} \citep{openai2025gpt5card}, which is not fine-tuned and therefore not a like-for-like comparison. 
The two encoder models fine-tuned in the multilingual setting are publicly available.\footnote{\url{https://hf.co/collections/DFKI-SLT/vestige}}

All experiments use a fixed document-level 70/15/15 split stratified by language. 
Models are evaluated using span-level partial-match precision (P), recall (R), and macro-$F_1$ computed with \texttt{nervaluate}.\footnote{\url{https://github.com/MantisAI/nervaluate}} 
Under this scheme, an exact boundary match receives full credit, while a prediction that partially overlaps a gold span receives half credit.
The multilingual test set consists of 77 documents, of which 26 are German, 26 English and 25 French (see results in \Cref{tab:compact-baselines_finalsplit}). 
Complete experimental details are provided in \Cref{app:baseline_details}. 

Both encoders detect PII reliably across languages ($F_1$ 0.83--0.93) and lose between 18 and 38 points on IPI, with the zero-shot \texttt{GPT-5.4} reference always lagging behind and English being the weakest language for all three systems.

\begin{table}[t]
  \centering
  \small
  \begin{tabular}{llccc}
    \toprule
    \textbf{Model} & \textbf{L} & \textbf{PII} & \textbf{IPI} & \textbf{All} \\
    \midrule
    \multirow{4}{*}{XLM-R}
      & Mix & 0.92 & 0.54 & 0.72 \\
      & DE  & 0.93 & 0.74 & 0.85 \\
      & EN  & 0.83 & 0.55 & 0.69 \\
      & FR  & 0.92 & 0.57 & 0.75 \\
    \midrule
    \multirow{4}{*}{mmBERT}
      & Mix & 0.91 & 0.54 & 0.72 \\
      & DE  & 0.91 & 0.72 & 0.83 \\
      & EN  & 0.84 & 0.53 & 0.69 \\
      & FR  & 0.93 & 0.60 & 0.77 \\
    \midrule
    \multirow{4}{*}{GPT-5.4}
      & Mix & 0.77 & 0.35 & 0.54 \\
      & DE  & 0.72 & 0.44 & 0.61 \\
      & EN  & 0.74 & 0.33 & 0.53 \\
      & FR  & 0.79 & 0.50 & 0.65 \\
    \bottomrule
  \end{tabular}
  \caption{Macro-averaged partial-match $F_1$ by identifier type present in the test set, across five training seeds. \texttt{GPT-5.4} was evaluated once per language setting. Precision, recall and standard deviations are in \Cref{app:baseline_details}; SD stays below 0.04 throughout. }
  \label{tab:compact-baselines_finalsplit}
\end{table}

\section{Auditing Residual Risk}
The annotation, risk taxonomy, and baselines converge in an interactive demonstrator that supports manual inspection of residual risk.
Given a system's predicted spans, it reports detection quality under a configurable span-matching criterion and 
links every missed gold identifier to the risk profile of its type.
Each document is summarized by the identifiers left exposed and its maximum residual risk, and flagged for manual review when a direct identifier was missed or at least three indirect ones remain.
Flagged documents can be inspected span by span with residual risk color-coded in context (\Cref{app:demo}), so a reviewer can decide whether to edit the document or refine the model.

Applying this to the mmBERT predictions, 43 of 77 evaluated documents ($\sim$56\%) were flagged, 12 of them containing a miss at the highest risk level, most often a named clinical facility.
A further 16 fell below the threshold but still retained undetected indirect identifiers, so some residual risk remains even where the automatic criteria do not trigger.
In the remaining 18, every gold identifier was detected, which reflects completeness with respect to our annotation scheme rather than a guarantee that no re-identifying information is left.
Fully detected documents were disproportionately short, formulaic patient requests (all languages), whereas flagged ones typically combined one missed direct identifier with several co-occurring indirect ones, 
which illustrates how combinations may compound into re-identification risk even when aggregate detection looks strong.

\section{Related Work}
\label{sec:related_work}

\paragraph{De-identification corpora.}
The resources named in the beginning differ in how they were built.
MEDDOCAN \citep{marimon_automatic_2019} and GRaSCCo \citep{modersohn_grascco_2022,lohr_-identifying_2024} are synthetic or alienated by construction, which is what makes them shareable, whereas i2b2/n2c2 \citep{stubbs_annotating_2015} contains de-identified real records and is currently unavailable\footnote{Listed as unavailable with no announced timeline as of July 24, 2026: \url{https://n2c2.dbmi.hms.harvard.edu/data-sets}}, illustrating how access to real-data benchmarks depends on institutional arrangements that can expire.
MultiGRaSCCo \citep{baroud_multigrascco_2026} extends GRaSCCo to ten languages by translation and supplies the indirect-identifier taxonomy we build on.
\citet{kovacevic_-identification_2024} survey the field and call for more manually labeled corpora across more domains, noting that reported identifier counts and preprocessing choices are often inconsistent.

\paragraph{Synthetic data for de-identification.}
\citet{deleger_preparing_2014} show that replacing HIPAA PHI with realistic surrogates preserves a corpus's research value, and later work moved from GANs \citep{kasthurirathne_generative_2021} to LLM pipelines.
\citet{shimizu_recordtwin_2025} $k$-anonymize extracted entities before prompting an LLM to fill the surrounding text, reaching near-original utility on MIMIC-III while reducing $n$-gram overlap with the source.
\citet{xie_differentially_2024} instead target formal, corpus-level privacy guarantees, on general-domain text and without identifier annotation.
None of these produce multilingual, multi-genre clinical text paired with an identifier-level risk taxonomy.

\paragraph{Re-identification risk frameworks.}
A few frameworks estimate risk directly rather than relying on detection recall.
\citet{scaiano_unified_2016} correct for identifier prevalence to estimate re-identification probability on one English corpus, and \citet{papadopoulou_neural_2026} score individual spans continuously using model probabilities, classifier confidence and web search, on English legal and biographical text.
RAT-Bench \citep{ratbench} evaluates multilingual anonymization tools on synthetic texts grounded in U.S.\ demographics, using an LLM attacker to estimate population-level risk.
Closest to our audit demonstrator, \citet{paul_deid-clinic_2026} add a document-level residual-risk module ($k$-anonymity, $l$-diversity, $t$-closeness) to a clinical pipeline for English discharge summaries.
Our dimensions follow the regulation-grounded scoring of \citet{prasser_importance_2016}, with uniqueness reflecting population-uniqueness models for structured data \citep{rocher_estimating_2019}.
None combine a  risk taxonomy with multilingual, multi-genre clinical text and an audit tool operating on system predictions.

\section{Discussion}
We found that seed language matters more than we expected.
English yields the lowest baseline scores for all three systems (\Cref{tab:compact-baselines_finalsplit}). All English documents were generated using French or German anchors, while German documents used French anchors for three genres. These anchors were supplied in their original language. The use of anchors in a different language from the generated document may have influenced the resulting texts.
Note, however, that our design cannot separate it from other language-specific factors, such as differences in identifier distributions (\Cref{tab:annot_stats}) or in model pretraining data.

Structural hallucination also remains a problem under blueprint control.
The model added dialogue-style passages and speaker labels in genres where the source data has none.
The invented material is clinically coherent but changes the surface form, so the pipeline works well where the source register is structured and less well where it is loose.
The small pilot study suggests that both intended audiences find the documents plausible, although with five raters this remains exploratory.

Annotation and detection show the gap the benchmark was built for (\Cref{tab:compact-baselines_finalsplit}).
PII forms a closed set with conventional surface forms and mostly clear boundaries, and is annotated and detected reliably.
IPI is harder on both counts: annotators disagree about where a span ends and about whether a passage is an identifier at all.
This is not an annotation failure but rather shows that what counts as identifying depends on the context and on the attacker, which we do not model here.
Risk scoring depends on context in the same way, but agreement was still moderate to good on all five dimensions.

Most identifier types reach an aggregated score of 3, for PII by our minimum legal-sensitivity rule and for IPI out of the scoring itself.
The maximum rule is conservative, so the aggregate is a screening signal rather than a ranking.
The audit makes this visible: an $F_1$ of 0.72 still leaves more than half of the documents flagged for review, and the per-document inspection lets a reviewer filter for the identifiers that are relevant in their own release scenario.

Extending \datasetname{} to further genres, such as hospital discharge summaries, requires a genre template with required and optional fields and at least one style anchor derived from de-identified seed documents.
The profile priors, blueprint schema and annotation guidelines can be reused as they are, and the guidelines already cover identifiers that are rare in our ambulatory data but likely frequent in inpatient genres, such as rare diagnoses. 

\section{Conclusion}
We introduced \datasetname, a multilingual benchmark for clinical de-identification and re-identification risk assessment, covering German, French, and English across eight healthcare genres used in production.
The two-stage pipeline produces synthetic documents with controllable style and content, and the annotation scheme provides span-level PII and IPI labels together with regulation-grounded risk scores per identifier type.
Baselines quantify how much harder indirect identifiers are to detect, and the audit tool links missed identifiers to their risk profiles to support manual review of residual risk.
Data, annotations, auditing tool and code can be found on Hugging Face\footnote{\url{https://hf.co/collections/doctolib-lab/vestige}, \url{https://hf.co/collections/DFKI-SLT/vestige}} and GitHub.\footnote{\url{https://github.com/DFKI-NLP/vestige}}

\section*{Limitations}
There are several limitations to the presented work.
The corpus is fully synthetic, so results obtained on it might not necessarily transfer to real production text.
Several genres in our data are generated automatically in deployment as well, which narrows the gap, but the patient-authored genres remain a synthetic approximation.
The seed data also comes from a single provider, so genre definitions, register, and identifier distributions reflect that provider's products.

With about 500 documents, the corpus is smaller than collections such as MIMIC.
The limiting factor is annotation effort rather than generation, and the pipeline can produce further documents as annotation capacity (but also resource availability) allows.
English documents were generated using French or German anchors, and documents in three German genres used French anchors. These anchors were supplied in their original language, which may have affected how closely the generated documents reflect the conventions of the target language and genre.
Genre frequencies are balanced artificially and depart from the production distribution.
This decision, however, was deliberate, to give every genre enough documents for evaluation and to diversify identifiers, but it should be taken into account.
The generated clinical content is mostly common, so rare-condition health identifiers, which carry the highest risk, are absent and cannot be evaluated with the current version of the corpus.

The human evaluation is a pilot with three patient and two clinician raters, so its results are exploratory.
Raters described the documents as almost ``too perfect'': they contain no typos, present medical information very clearly, and over-state some attributes, e.g., explicitly mentioning that a patient does not smoke, likely because sampled profile attributes are verbalized.
Real documents are noisier, which may make detection on \datasetname{} easier than on production text.

Agreement on indirect identifiers is moderate ($\alpha=0.584$), which restricts how precisely systems can be evaluated on that aspect.
The current test set comprises only 77 documents, so small differences between the encoders should not be over-interpreted.
Risk scores are assigned per identifier type rather than per mention, so a rare and a common surname carry the same score despite different uniqueness in context.
Importantly, the risk scores are expert judgments and have not been validated against actual linkage attacks, so they serve as an inspection aid rather than as estimates of re-identification probability.
The document-level flagging rule follows an established convention but its thresholds are not empirically tuned on our data.
Finally, the corpus covers ambulatory care and may not generalize to hospital discharge summaries or other settings without adaptation.

\section*{Ethical Considerations}
 
All documents in this corpus are fully synthetic and contain no real patient data.
The seed documents came from a European digital-health provider.
Most genres were produced in-house (``actors'') or synthesized by the provider for production testing.
The remainder were anonymized and manually reviewed on-premise before transfer, in a multi-step process carried out by the provider and agreed upon as a condition of the collaboration.
\textit{No original document left the provider's environment}, and the style anchors used during generation were derived from reviewed seed documents rather than copied from them, so no source document was transmitted to a third-party model provider.
Seed documents will be deleted upon completion of the collaboration.

Person and facility names were generated by the LLM, with repeated names replaced from curated country-specific name pools (\Cref{sec:pipeline}), and were checked to avoid resemblance to real individuals or institutions.
Use of the dataset for training and evaluating de-identification systems is encouraged.
Use for re-identification of real individuals is prohibited, and this restriction will be part of the release license, as for other de-identification datasets.

Our taxonomy and risk profiles describe which features make clinical text re-identifiable, which is information an attacker could also use.
We consider publication justified because the same information is required to build and audit protective systems, and because the underlying categories are already documented in data protection guidance.
The audit tool likewise reports residual risk relative to our annotation scheme and thresholds. If a document is not flagged, it means that it is below those thresholds, but it does not certify it as anonymous, and the tool is intended to assist human review rather than replace it.

The clinical content of the documents is invented.
It should not be used to train or evaluate systems that give clinical advice.
Annotators were employed at our institute, with student assistants compensated at standard institutional rates.
The participants of the human evaluation took part voluntarily after being informed about the purpose of the study, and no personal data about them were collected.

\section*{Acknowledgments}
We gratefully acknowledge funding from the German Federal Ministry of Research, Technology and Space (BMFTR) through the project VERANDA (16KIS2046K) and through the grant BIFOLD26B.
This work was co-funded by the European Commission under grant agreement No.\ 101290710 (help.D).
The authors also thank the participants of the human evaluation and our annotators.

\newpage
\bibliography{custom}

\appendix

\section{Details on generated documents}
\label[appendix]{app:details_generated_docs}

\begin{table}[H]
  \centering
  \footnotesize
  \begin{tabular}{lrrr}
    \toprule
    \textbf{Genre} & \textbf{DE} & \textbf{EN} & \textbf{FR} \\
    \midrule
    consult.-asst. dial.       & 18 & 18 & 18 \\
    consult.-asst. sum.        & 16 & 16 & 16 \\
    consult.-asst. trans.     & 10 & 10 &  9 \\
    referral letters                 & 10 & 10 &  9 \\
    medical dictation                 & 26 & 26 & 26 \\
    patient request                   & 54 & 54 & 54 \\
    phone assistant                   & 16 & 16 & 16 \\
    AIHA                & 20 & 20 & 20 \\
    \midrule
    \textbf{Total}                    & \textbf{170} & \textbf{170} & \textbf{168} \\
    \bottomrule
  \end{tabular}
  \caption{Final document counts by genre and language. Two French documents did not pass the quality tests.}
  \label{tab:genredist}
\end{table}

\begin{table}[H]
  \centering
  \footnotesize
  \setlength{\tabcolsep}{4pt}
  \begin{tabular}{llrr}
    \toprule
    \textbf{L} & \textbf{Genre} &
    \textbf{$\overline{cos}_\text{true}$} &
    \textbf{$\overline{cos}_\text{rand}$} \\
        \midrule
    DE & consult.-asst. dial. & 0.436 & 0.359 \\
    DE & consult.-asst. sum. & 0.592 & 0.302 \\
    DE & consult.-asst trans. & 0.442 & 0.378 \\
    DE & referral letters & 0.673 & 0.316 \\
    DE & medical dictation & 0.626 & 0.345 \\
    DE & patient request & 0.404 & 0.349 \\
    DE & phone assistant & 0.719 & 0.361 \\
    DE & AIHA & 0.171 & 0.243 \\
    \midrule
    EN & consult.-asst. dial. & 0.479 & 0.390 \\
    EN & consult.-asst sum. & 0.713 & 0.335 \\
    EN & consult.-asst-trans. & 0.523 & 0.377 \\
    EN & referral letters & 0.631 & 0.300 \\
    EN & medical dictation & 0.674 & 0.343 \\
    EN & patient request & 0.430 & 0.358 \\
    EN & phone assistant & 0.669 & 0.394 \\
    EN & AIHA & 0.258 & 0.225 \\
    \midrule
    FR & consult.-asst. dial. & 0.558 & 0.365 \\
    FR & consult.-asst. sum. & 0.630 & 0.350 \\
    FR & consult.-asst. trans. & 0.481 & 0.393 \\
    FR & referral letters & 0.649 & 0.352 \\
    FR & medical dictation & 0.446 & 0.397 \\
    FR & patient request & 0.575 & 0.339 \\
    FR & phone assistant & 0.759 & 0.410 \\
    FR & AIHA & 0.104 & 0.224 \\
    \bottomrule
  \end{tabular}
  \caption{Mean cosine similarity of synthetic documents to their true anchor vs.\ unrelated sampled anchor documents, computed with \texttt{jinaai/jina-embeddings-v3} and 8192 tokens sequence length. 
  }
  \label{tab:cosine}
\end{table}

\begin{table}[H]
  \centering
  \footnotesize
  \setlength{\tabcolsep}{4pt}
  \begin{tabular}{llrr}
    \toprule
    \textbf{L} & \textbf{Genre} &
    \textbf{$\overline{\mathrm{BScore}}_{\mathrm{true}}$} &
    \textbf{$\overline{\mathrm{BScore}}_{\mathrm{rnd}}$} \\
    \midrule
    DE & consult.-asst. dial. & 0.827 & 0.809 \\
    DE & consult.-asst sum. & 0.828 & 0.806 \\
    DE & consult.-asst trans. & 0.812 & 0.794 \\
    DE & referral letters & 0.815 & 0.787 \\
    DE & medical dictation & 0.831 & 0.806 \\
    DE & patient request & 0.864 & 0.807 \\
    DE & phone assistant & 0.859 & 0.810 \\
    DE & AIHA           & 0.828 & 0.794 \\
    \midrule
    EN & consult.-asst. dial. & 0.818 & 0.803 \\
    EN & consult.-asst. sum. & 0.821 & 0.799 \\
    EN & consult.-asst. trans. & 0.815 & 0.791 \\
    EN & referral letters & 0.817 & 0.792 \\
    EN & medical dictation & 0.810 & 0.801 \\
    EN & patient request & 0.850 & 0.795 \\
    EN & phone assistant & 0.850 & 0.799 \\
    EN & AIHA            & 0.828 & 0.802 \\
    \midrule
    FR & consult.-asst. dial. & 0.839 & 0.808 \\
    FR & consult.-asst. sum. & 0.820 & 0.807 \\
    FR & consult.-asst. trans. & 0.835 & 0.806 \\
    FR & referral letters & 0.840 & 0.805 \\
    FR & medical dictation & 0.835 & 0.802 \\
    FR & patient request & 0.890 & 0.806 \\
    FR & phone assistant & 0.874 & 0.811 \\
    FR & AIHA            & 0.826 & 0.824 \\
    \bottomrule
  \end{tabular}
  \caption{Mean BERTScore from true anchor vs. random anchor.}
  \label{tab:bertscore_crossgenre}
\end{table}

\begin{table}[H]
  \centering
  \footnotesize
  \begin{tabular}{llrr}
    \toprule
    \textbf{L} & \textbf{Genre} & \textbf{Word ratio} &
    \textbf{Mean tokens} \\
    \midrule
    DE & consult.-asst. dial.   & 0.51 & 435.83 \\
    DE & consult.-asst. sum.         & 1.76 & 253.94 \\
    DE & medical-dictation         & 1.00 & 488.81 \\
    DE & patient-request           & 1.54 & 42.56  \\
    DE & phone-assistant           & 1.54 & 474.06 \\
    \midrule
    FR & consult.-asst. dial.   & 0.26 & 366.72 \\
    FR & consult.-asst. trans. & 1.15 & 805.33 \\
    FR & consult.-asst. sum.     & 1.74 & 302.06 \\
    FR & referral letters         & 1.66 & 297.33 \\
    FR & medical-dictation         & 1.30 & 490.31 \\
    FR & patient-request           & 0.83 & 45.41  \\
    FR & phone-assistant           & 1.46 & 493.56 \\
    FR & AIHA                      & 7.81 & 62.95  \\
    \bottomrule
  \end{tabular}
  \caption{Generated-to-seed word-count ratios and mean
  synthetic-document token counts, by language and genre.
  Rows are shown only where seed documents exist in the target language. 
  There are no English seed documents, and no German ones for \textit{consultation-assistant transcripts}, \textit{referral letters} and \textit{AIHA requests}.}
  \label{tab:lengths}
\end{table}

\subsection{Annotators (Identifiers)}
\label[appendix]{app:anno_backgrounds}
In total, four annotators contributed to the task of annotating identifiers in three languages.
They have complementary backgrounds in business informatics, medicine, linguistics, and computational linguistics. 
All are native speakers of German and were employed at our institute.
Three annotators were student assistants compensated at standard rates, and the fourth is regular staff. 
Two are highly experienced annotators, with over five years of annotation practice respectively, and have authored several annotation guidelines, including guidelines for de-identification of texts, the other two have two or more years of annotation experience. 
One annotator has a medical background (second state examination in Germany) and prior clinical work experience in France. 
Language coverage spans all three dataset languages: all four annotators are proficient in German and English, two are fluent or highly proficient in French (having studied or worked in France), and one annotator has school-level French. 
Two annotators worked on English and German data, one on English and French, and one on all three languages.

\subsection{IAA for Identifier Annotation}
\label[appendix]{app:iaa_identifiers}

\begin{table}[H]
    \centering
    \small
    \begin{tabular}{lcc}
        \toprule
        \textbf{Metric} & \textbf{PII} & \textbf{IPI} \\
        \midrule
        Krippendorff's $\alpha_{\text{unitizing}}$ & 0.907 & 0.584 \\
        macro $F_1$ exact & 0.797 & 0.384 \\
        macro $F_1$ relaxed & 0.927 & 0.669 \\
        micro $F_1$ exact & 0.834 & 0.358 \\
        micro $F_1$ relaxed & 0.928 & 0.675 \\
        \bottomrule
    \end{tabular}
    \caption{Overall inter-annotator agreement for directly (PII) and indirectly identifying information (IPI).}
    \label{tab:iaa_overall}
\end{table}

\begin{table}[H]
  \centering
  \small
  \setlength{\tabcolsep}{4pt}
  \begin{tabular}{@{}llccc@{}}
    \toprule
    \textbf{Metric} & & \textbf{DE} & \textbf{EN} & \textbf{FR} \\
    \midrule
    \multirow{2}{*}{$\alpha_\text{unitizing}$}
      & PII & 0.902 & 0.899 & \textbf{0.921} \\
      & IPI & \textbf{0.614} & 0.572 & 0.574 \\
    \addlinespace[2pt]
    \multirow{2}{*}{macro $F_1$}
      & PII & 0.778 & 0.784 & \textbf{0.829} \\
      & IPI & \textbf{0.428} & 0.362 & 0.362 \\
    \addlinespace[2pt]
    \multirow{2}{*}{micro $F_1$}
      & PII & 0.826 & \textbf{0.845} & 0.823 \\
      & IPI & \textbf{0.384} & 0.360 & 0.324 \\
    \addlinespace[2pt]
    \multirow{2}{*}{macro $F_1$ rel.}
      & PII & 0.925 & 0.912 & \textbf{0.942} \\
      & IPI & 0.622 & \textbf{0.700} & 0.684 \\
    \addlinespace[2pt]
    \multirow{2}{*}{micro $F_1$ rel.}
      & PII & \textbf{0.934} & 0.929 & 0.921 \\
      & IPI & 0.617 & \textbf{0.701} & 0.691 \\
    \bottomrule
  \end{tabular}
  \caption{Inter-annotator agreement by language. Bold marks the highest value per row. Relaxed $F_1$ counts partially overlapping spans as matches.}
  \label{tab:iaa_language}
\end{table}

\begin{table}[H]
  \centering
  \small
  \begin{tabular}{lrr}
    \toprule
    \textbf{Identifier-level agreement} & \textbf{$N$} & \textbf{$\alpha_u$} \\
    \midrule

    \multicolumn{3}{l}{\textbf{PII}} \\
    \textsc{Name\_Client}        & 115 & 0.926  \\
    \textsc{Name\_Professional}  & 364 & 0.986  \\
    \textsc{Name\_Social}        & 5   & 0.155  \\
    \textsc{Date}                & 547 & 0.893  \\
    \textsc{Age}                 & 92  & 0.614  \\
    \textsc{Geolocation}         & 596 & 0.633  \\
    \textsc{Facility}            & 306 & 0.958  \\
    \textsc{Contact\_Phone}      & 64  & 0.930  \\
    \textsc{Profession}          & 89  & 0.080  \\

    \midrule
    \multicolumn{3}{l}{\textbf{IPI}} \\
    \textsc{Appearance}           & 2   & $-0.028$ \\
    \textsc{Time}                 & 608 & 0.685  \\
    \textsc{Age}                  & 230 & 0.531  \\
    \textsc{Geolocation}          & 7   & 0.134  \\
    \textsc{Facility}             & 39  & $-0.012$ \\
    \textsc{Lifestyle}            & 394 & 0.603  \\
    \textsc{Context}              & 24  & $-0.085$ \\
    \textsc{Demographics}         & 19  & $-0.012$ \\
    \textsc{Dem.\_Family\_Household}    & 171 & 0.722  \\
    \textsc{Dem.\_Socio}                & 473 & 0.550  \\
    \textsc{Quantity}             & 135 & 0.002  \\
    \bottomrule
  \end{tabular}

  \caption{
    Identifier-level IAA (unitizing Krippendorff's $\alpha_u$) before curation.
$N$: document--annotator pairs.
\textsc{Name} subcategories were later collapsed into \textsc{Name}, and IPI \textsc{Facility}, \textsc{Demographics} and \textsc{Appearance} were merged during curation (cf.\ \Cref{tab:annot_stats}).
  }
  \label{tab:iaa_identifier}
\end{table}

\section{Risk Auditing}

\subsection{Curation and Risk Profiles}
\label[appendix]{app:risk_profiles}

\paragraph{Curation}
\Cref{tab:curated_risk} shows the curation results of the risk assessment per identifier type. 
Exact agreement as well as one-step disagreement happened often, while only once an adjudication was necessary when annotators disagreed by two points on the scale. 
Minor disagreements ($|\Delta|=1$, 27.9\%) were resolved by taking the higher score, consistent with the privacy-protective purpose of the benchmark and the assumption that the annotator assigning the higher value may have identified a linkage vector or contextual risk the other did not. 
Stability accounts for the largest share of these resolutions (39.3\%), consistent with its comparatively lower $\alpha_{risk}$ (0.609) and the general difficulty of judging how persistently an attribute identifies a person over time. 
The single two-step disagreement ($|\Delta|=2$, 0.7\%) concerned \textsc{Facility} (IPI) on legal sensitivity: one annotator rated the risk low (1), reasoning that a generically named healthcare facility rarely reveals a diagnosis, while the other rated it high (3), noting that some facility types (e.g., psychiatric or oncology clinics) can implicitly reveal a diagnosis or treatment type under GDPR Article 9. 
Curation settled on the middle value (2).

\begin{table}[H]
\centering
\small
\begin{tabular}{lrrrr}
\toprule
Dimension & Exact & Max\,($\uparrow$) & Adj. & Total \\
\midrule
Membership        & 23 (82)  &  5 (18)  & 0 (0)  & 28 \\
Uniqueness        & 19 (68)  &  9 (32)  & 0 (0)  & 28 \\
Stability         & 17 (61)  & 11 (39)  & 0 (0)  & 28 \\
Linkability       & 21 (75)  &  7 (25)  & 0 (0)  & 28 \\
Legal             & 20 (71)  &  7 (25)  & 1 (4)  & 28 \\
\midrule
\textbf{Total}    & \textbf{100 (71)} & \textbf{39 (28)} & \textbf{1 (1)} & \textbf{140} \\
\bottomrule
\end{tabular}
\caption{Annotation curation breakdown per risk dimension ($n$=28 identifier types). 
         Numbers show count (percentage). 
         \emph{Exact} ($|\Delta|=0$): both annotators assign the same score, and the values are retained directly.
         \emph{Max\,($\uparrow$)} ($|\Delta|=1$): one-step disagreement, resolved to higher score.
         \emph{Adjudicate} (Adj.) ($|\Delta|=2$): maximum possible disagreement on the three-point scale, these cases (1) were manually curated.}
\label{tab:curated_risk}
\end{table}

\paragraph{IAA for risk assessment}
\Cref{tab:iaa_risk_detailed} shows a detailed overview of inter-annotator agreement per identifier type and dimension.
\Cref{tab:gold_risk_assessment} presents the final risk profiles after curation.

\paragraph{Prevalence bias in agreement}
For PII, membership produced a near-zero $\alpha_{risk}$ of $-$0.050 despite an exact agreement rate of 81.8\% (marked with $^\dagger$ in \Cref{tab:iaa_risk_detailed}). 
This is a known consequence of prevalence bias in chance-corrected metrics: PII identifiers by definition disclose cohort membership and fall under GDPR Article~9 special categories, causing both annotators to assign ceiling scores consistently. 
When score distributions are strongly skewed, expected chance agreement approaches observed agreement, rendering $\alpha_{risk}$ uninformative rather than indicative of poor annotation quality.
Therefore, exact agreement is the more appropriate summary for this dimension on PII.

A milder version of the same pattern appears for legal sensitivity on PII ($\alpha_{risk}=0.259$, exact agreement 72.7\%): every PII type is by definition explicitly named under GDPR or HIPAA, again limiting scale variance. IPI legal sensitivity, whose status is more heterogeneous across types, shows substantial agreement ($\alpha_{risk}=0.629$).

\paragraph{Pooling effect on overall values}
For membership and linkability, overall $\alpha_{risk}$ (0.852, 0.742) exceeds both the PII (\,$-$0.050, 0.417\,) and IPI (0.731, 0.425) values. 
This is not an averaging artifact: overall is computed on the pooled reliability matrix of all 28 identifier types, and Krippendorff's alpha corrects for chance using the marginal score distribution of the group being evaluated. 
Pooling two subgroups with different marginals shifts that chance baseline in non-additive ways, so the pooled estimate need not fall between the subgroup values. 
This is a known behavior of chance-corrected agreement statistics, closely related to the prevalence paradox above \citep{feinstein_high_1990,cicchetti_high_1990}.

\begin{table}[H]
\centering
\small
\setlength{\tabcolsep}{4pt}
\begin{tabular}{@{}lccc@{}}
\toprule
Dimension & PII ($n$=11) & IPI ($n$=17) & All ($n$=28) \\
\midrule
Membership   & $-$0.05$^\dagger$/0.00 & 0.731/0.712 & 0.852/0.767 \\
Uniqueness   & 0.462/0.441            & 0.709/0.579 & 0.667/0.590 \\
Stability    & 0.494/0.371            & 0.576/0.425 & 0.609/0.442 \\
Linkability  & 0.417/0.421            & 0.425/0.361 & 0.742/0.623 \\
Legal sens.  & 0.259$^\dagger$/0.298  & 0.629/0.605 & 0.579/0.561 \\
\bottomrule
\end{tabular}
\caption{Inter-annotator agreement per risk dimension, reported as $\alpha_{risk}$/$\kappa_{risk}$ (Krippendorff's alpha with ordinal distance / linearly weighted Cohen's kappa). Interpretation follows \citet{landis_measurement_1977}. $^\dagger$Prevalence bias is likely: chance-corrected metrics are uninformative when score distributions are strongly skewed
(exact agreement: membership 81.8\%, legal sensitivity 72.7\%).}
\label{tab:iaa_risk_detailed}
\end{table}

\begin{table*}[t]
  \centering
  \begin{tabular}{lrrrrrr}
    \toprule
    \textbf{Identifier} & \textbf{Mem} & \textbf{Uni} & \textbf{Sta} & \textbf{Lnk} & \textbf{Leg} & \textbf{Agg} \\
    \midrule
    \multicolumn{7}{l}{\textit{Personally Identifiable Information (PII)}} \\
    \textsc{Name}                        & 3 & 2 & 3 & 3 & 3 & 3 \\
    \textsc{Date}                        & 3 & 3 & 3 & 3 & 3 & 3 \\
    \textsc{Age}                         & 3 & 3 & 2 & 3 & 3 & 3 \\
    \textsc{Geolocation}                 & 3 & 3 & 2 & 3 & 3 & 3 \\
    \textsc{Facility}                    & 3 & 2 & 3 & 3 & 3 & 3 \\
    \textsc{Organization}                & 3 & 3 & 3 & 3 & 3 & 3 \\
    \textsc{Contact\_Phone}              & 3 & 3 & 2 & 3 & 3 & 3 \\
    \textsc{Contact\_Online}             & 3 & 3 & 3 & 3 & 3 & 3 \\
    \textsc{Code}                        & 3 & 3 & 3 & 2 & 3 & 3 \\
    \textsc{Biometric}                   & 3 & 3 & 3 & 3 & 3 & 3 \\
    \textsc{Profession}                  & 3 & 2 & 3 & 3 & 2 & 3 \\
    \midrule
    \multicolumn{7}{l}{\textit{Indirect Personal Information (IPI)}} \\
    \textsc{Appearance}                  & 3 & 3 & 3 & 3 & 2 & 3 \\
    \textsc{Time}                        & 1 & 2 & 2 & 2 & 1 & 2 \\
    \textsc{Age}                         & 2 & 1 & 2 & 2 & 1 & 2 \\
    \textsc{Geolocation}                 & 2 & 1 & 3 & 2 & 3 & 3 \\
    \textsc{Facility}                    & 2 & 2 & 3 & 2 & 2 & 3 \\
    \textsc{Disease\_Diagnosis\_Procedure} & 3 & 3 & 3 & 3 & 3 & 3 \\
    \textsc{Symptoms}            & 2 & 2 & 2 & 2 & 3 & 3 \\
    \textsc{Medication}          & 2 & 3 & 2 & 2 & 3 & 3 \\
    \textsc{Result}              & 2 & 3 & 2 & 2 & 3 & 3 \\
    \textsc{History}             & 2 & 3 & 2 & 2 & 3 & 3 \\
    \textsc{Legal}                       & 2 & 3 & 3 & 2 & 3 & 3 \\
    \textsc{Lifestyle}                   & 2 & 2 & 2 & 2 & 1 & 2 \\
    \textsc{Context}                     & 2 & 3 & 2 & 2 & 2 & 3 \\
    \textsc{Demographics}                & 2 & 2 & 3 & 2 & 3 & 3 \\
    \textsc{Dem.\_Family\_household}           & 1 & 1 & 2 & 2 & 2 & 2 \\
    \textsc{Dem.\_Socio}                       & 2 & 2 & 2 & 2 & 3 & 3 \\
    \textsc{Quantity}                    & 1 & 1 & 2 & 1 & 2 & 2 \\
    \bottomrule
  \end{tabular}
  \caption{
    Risk profiles: Assessment of identifiers with respect to residual re-identification risk.
    Scores are assigned per identifier type on five dimensions (1 = low risk, 3 = high risk).
    Mem = membership, Uni = uniqueness, Sta = stability, Lnk = linkability, Leg = legal sensitivity, Agg = aggregated risk score (maximum over five dimensions).
  }
  \label{tab:gold_risk_assessment}
\end{table*}

\subsection{Auditing Risk (Demo)}
\begin{figure*}[h]
    \centering
    \includegraphics[width=0.95\linewidth]{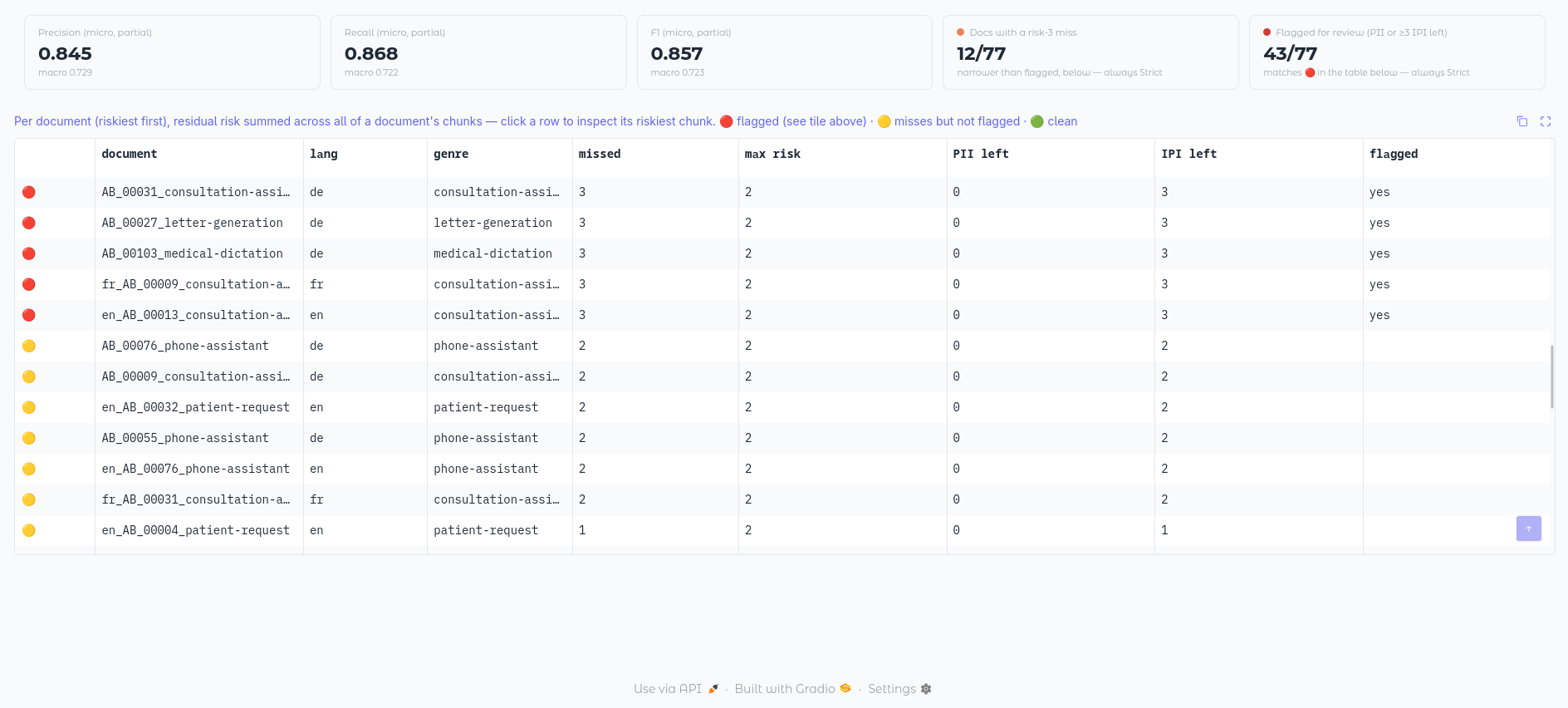}
    \caption{Evaluation view of the audit tool: Upload predictions in the test data and get one risk assessment per document as well as the overall evaluation scores (micro scores in bold, macro in grey below). }
    \label{fig:evaluation}
\end{figure*}

\begin{figure*}[h]
    \centering
    \includegraphics[width=0.95\linewidth]{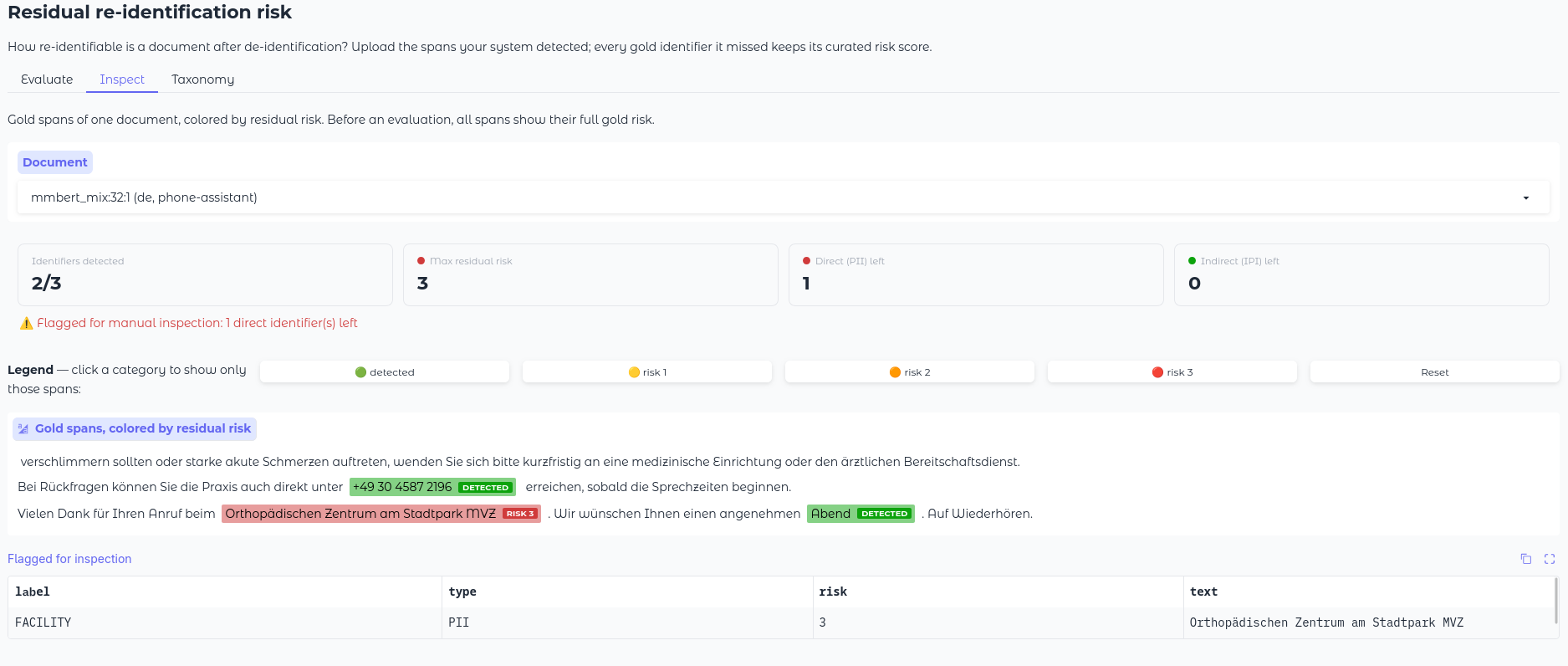}
    \caption{Inspection view of the demonstrator (chunk of document shown): gold identifiers of one document, colored by residual risk (green = detected; yellow/orange/red = missed, risk 1-3). 
    Document-level tiles summarize detected identifiers, maximum residual risk, and remaining direct/indirect identifiers. 
    Documents are flagged for manual review when a direct identifier is missed or $\geq$ 3 indirect identifiers remain.}
    \label{fig:inspection}
\end{figure*}

\label[appendix]{app:demo}
The audit tool is presented as a demo.
\Cref{fig:evaluation} shows the evaluation overview.
A second tab allows the user to inspect each document separately, displaying the detected and non-detected identifiers and their accompanying risk (\Cref{fig:inspection}). 

\section{Details of the human evaluation}
\label[appendix]{app:user_study_details}

The exact questions asked in the pilot study for patients and clinicians are shown in \Cref{tab:full-questionnaires}.

\begin{table}[H]
\centering
\footnotesize
\begin{tabular}{@{}p{0.7cm}p{6.4cm}@{}}
\toprule
\textbf{ID} & \textbf{Question} \\
\midrule
\multicolumn{2}{@{}l}{\textbf{Clinician questionnaire}} \\
C1 & How realistic does this document appear as clinical documentation? \\
C2 & How internally consistent is the clinical information presented in this document? \\
C3 & To what extent does this document provide sufficient information to formulate an appropriate next clinical step without additional clarification? \\
C4 & To what extent does this document appear complete and free of information gaps for appropriate follow-up? \\
C5 & Overall, how suitable would this document be for use in routine clinical practice? \\
C6 & Comment (free-text response). \\
\midrule
\multicolumn{2}{@{}l}{\textbf{Patient questionnaire}} \\
P1 & How well would you understand what is happening and why after reading this document? \\
P2 & How well would you know what to do next after reading this document? \\
P3 & To what extent would this document make you feel that your concerns were understood? \\
P4 & How trustworthy does the information in this document seem to you? \\
P5 & Overall, how helpful would this document be for understanding your healthcare? \\
P6 & Comments (free-text response). \\
\bottomrule
\end{tabular}
\caption{Complete clinician and patient questionnaires. Identifiers correspond to the abbreviated references used in the main paper.}
\label{tab:full-questionnaires}
\end{table}

\section{Prompt Template \& Generation Hyperparameters}
\label[appendix]{app:prompt_template}

For each synthetic document, the pipeline uses a two-stage prompting procedure. First, a blueprint-generation prompt is used to create a structured ``intermediate'' representation. Second, a rendering prompt is used to generate the final free-text document from that blueprint.

Both LLM calls use \texttt{GPT-5.3-chat} \cite{openai2025gpt5card} through Microsoft Azure. This safeguards that prompts and outputs are not used for model training. The model is configured with a temperature of 0.0, top-$p$ of 1.0, and a maximum output length of 1,800 tokens.

\begin{figure*}[t]
\centering
\begin{tcolorbox}[
    width=0.97\linewidth,
    colback=orange!5!white,
    colframe=gray!75!black,
    title=Blueprint Generation Prompt Template
]
\small

\textbf{System Message}\\

You create medical document blueprints for synthetic data generation.  
Return JSON only. It should read like ordinary real-world data.  
Do not copy the anchor text.\\

\vspace{0.5em}
\textbf{User Message}\\

Create one JSON blueprint for a synthetic clinical document.\\

\texttt{doc\_id: <DOC\_ID>}\\
\texttt{genre: <GENRE>}\\
\texttt{target\_language: <LANG>}\\
\texttt{target\_token\_range: <TOKEN\_MIN>..<TOKEN\_MAX>}\\
\texttt{sampled\_profile\_controls: <SAMPLED\_PROFILE\_CONTROLS>}\\
\texttt{required\_schema\_slots: <REQUIRED\_SCHEMA\_SLOTS>}\\
\texttt{optional\_schema\_slots: <OPTIONAL\_SCHEMA\_SLOTS>}\\
\texttt{anchor\_path: <ANCHOR\_PATH>}\\
\texttt{anchor\_language: <ANCHOR\_LANGUAGE>}\\

For \texttt{<LANG>} output, use names, institutions, addresses, dates, abbreviations, and clinical phrasing that look native to \texttt{<LANG>}-speaking medical and administrative usage. If person or organization fields are present, prefer ordinary contemporary \texttt{<LANG>}-style names and plausible clinic or hospital naming patterns.\\

Use exactly this one anchor for high-level style, formatting, rhythm, and genre cues only.  
Do not reuse names, dates, addresses, phone numbers, clinics, or sentences from it.  
When filling schema slots, prefer culturally and linguistically plausible values for the target language.  
Names of patients, doctors, clinics, hospitals, and cities should feel ordinary and native for that language.  
They must not sound cartoonish, placeholder-like, exaggerated, self-consciously synthetic, or obviously machine-made.\\

Do not use labels or markers such as fictive, fictional, fake, dummy, mock, test, sample, synthetic, invented, placeholder, example hospital, or example patient inside the slot values.  
Do not use redaction placeholders or angle-bracket markers such as \texttt{<FIRST\_NAME>}, \texttt{<LAST\_NAME>}, \texttt{<DATE>}, or \texttt{<CITY>}.  
Invent new values, but make them subtle, naturalistic, and unremarkable.\\

Return a JSON object with exactly these top-level keys: doc\_id, genre, lang, anchor\_path, anchor\_language, sampled\_profile\_controls, schema\_payload, blueprint\\

Constraints:
\begin{itemize}\setlength\itemsep{0.2em}
\item \texttt{schema\_payload} must contain all required schema slots and may contain optional schema slots.
\item No keys outside the required/optional schema slot lists may be introduced.
\item \texttt{blueprint} must contain:
\begin{itemize}\setlength\itemsep{0.2em}
\item \texttt{visit\_context}
\item \texttt{chief\_complaint}
\item \texttt{key\_clinical\_facts} (list of strings)
\item \texttt{diagnostic\_or\_operational\_impression}
\item \texttt{care\_plan} (list of strings)
\item \texttt{style\_notes} (list of strings)
\end{itemize}
\end{itemize}

Anchor text follows: \texttt{<ANCHOR\_TEXT>}

\end{tcolorbox}
\caption{Prompt template used for blueprint generation in the anchor-based synthetic document pipeline. The model is instructed to return a structured JSON object containing demographic controls, genre-specific schema content, and a semantic outline.}
\label{fig:blueprint_generation_prompt}
\end{figure*}

\begin{figure*}[t]
\centering
\begin{tcolorbox}[
    width=0.97\linewidth,
    colback=orange!5!white,
    colframe=gray!75!black,
    title=Document Rendering Prompt Template
]
\small

\textbf{System Message}\\

You write medical documents for synthetic data generation.  
Return plain text only. Do not mention that the document is synthetic.  
The result should read like an ordinary in-domain document written by a native user of the target language.\\

\vspace{0.5em}
\textbf{User Message}\\

Write one document from the provided blueprint.\\

\texttt{genre: <GENRE>}\\
\texttt{target\_language: <LANG>}\\
\texttt{target\_token\_range: <TOKEN\_MIN>..<TOKEN\_MAX>}\\
\texttt{anchor\_path: <ANCHOR\_PATH>}\\
\texttt{anchor\_language: <ANCHOR\_LANGUAGE>}\\

For \texttt{<LANG>} output, use names, institutions, addresses, dates, abbreviations, and clinical phrasing that look native to \texttt{<LANG>}-speaking medical and administrative usage. If person or organization fields are present, prefer ordinary contemporary \texttt{<LANG>}-style names and plausible clinic or hospital naming patterns.\\

Use the anchor only for style, register, structure, and rhythm.  
Do not copy wording, named entities, dates, numbers, addresses, or phone numbers.  
If the anchor language differs from the target language, transfer only the genre structure and pacing.  
Use the blueprint semantics faithfully. Keep profile controls editable by reflecting the sampled values naturally when the genre supports it, but do not force awkward demographic exposition.\\

If the schema includes names of patients, doctors, clinics, hospitals, or cities, render them as natural, plausible entities for the target language and setting.  
They should feel realistic and routine, not theatrical, playful, obviously fabricated, or visibly synthetic.  
Do not insert labels such as fictional, fake, dummy, mock, synthetic, invented, placeholder, or example into the text.  
Do not output redaction placeholders or angle-bracket markers such as \texttt{<FIRST\_NAME>}, \texttt{<LAST\_NAME>}, \texttt{<DATE>}, or \texttt{<CITY>}.\\

Blueprint and anchor are provided as: \texttt{<BLUEPRINT\_JSON>}\\

Anchor text follows: \texttt{<ANCHOR\_TEXT>}\\

Output text only.

\end{tcolorbox}
\caption{Prompt template used for document rendering in the anchor-based synthetic document pipeline. The model receives the complete validated blueprint together with the anchor text and generates the final free-text document.}
\label{fig:document_rendering_prompt}
\end{figure*}

\section{Baseline Details}
\label[appendix]{app:baseline_details}

We tuned each combination of model and training corpus (German, English, French and multilingual) independently using a full grid search. The grid consisted of learning rates (2e-5, 3e-5, 5e-5), epochs (8, 12) and two values for the per-device batch size and gradient-accumulation steps. They were (2, 4) and (1, 2) respectively. This resulted in 24 configurations per model--corpus combination. We fixed the maximum context to 512 tokens, weight decay to 0.01 and a warm-up ratio of 0.06. We selected the configuration with the highest validation macro-$F_1$ and applied early stopping with a patience of three evaluations and a minimum improvement of 0.001. The selected configuration was then retrained on the union of training and validation sets using five seeds. Consequently, each reported encoder baseline aggregates five independent final runs. Values are reported as the mean and sample standard deviation. In total, the baseline experiment comprised 192 tuning runs and 40 final runs on one NVIDIA RTX A6000 GPU.

The \texttt{GPT-5.4} baseline was evaluated using zero-shot prompting in a single run. The model was given task instructions together with brief descriptions of the direct and indirect identifier categories and was asked to identify all spans in each document. Predictions were evaluated using the same span-level partial-match metric as the encoder models.

\begin{table*}[!h]
  \centering
  \footnotesize
  \begin{tabular}{lllccc}
    \toprule
    \textbf{System} & \textbf{Lang} & \textbf{Category}
      & \textbf{P} & \textbf{R} & \textbf{F1} \\
    \midrule

    \multirow{12}{*}{XLM-R-large}
      & Mixed & PII     & 0.945 \sd{0.021} & 0.911 \sd{0.003} & 0.924 \sd{0.007} \\
      &       & IPI     & 0.518 \sd{0.013} & 0.567 \sd{0.009} & 0.540 \sd{0.008} \\
      &       & Overall & 0.717 \sd{0.016} & 0.728 \sd{0.005} & 0.719 \sd{0.007} \\
      \cmidrule(lr){2-6}
      & DE    & PII     & 0.951 \sd{0.010} & 0.905 \sd{0.006} & 0.926 \sd{0.006} \\
      &       & IPI     & 0.728 \sd{0.022} & 0.773 \sd{0.019} & 0.742 \sd{0.020} \\
      &       & Overall & 0.859 \sd{0.010} & 0.850 \sd{0.010} & 0.850 \sd{0.011} \\
      \cmidrule(lr){2-6}
      & EN    & PII     & 0.828 \sd{0.006} & 0.835 \sd{0.007} & 0.831 \sd{0.006} \\
      &       & IPI     & 0.534 \sd{0.005} & 0.581 \sd{0.004} & 0.554 \sd{0.004} \\
      &       & Overall & 0.681 \sd{0.004} & 0.708 \sd{0.003} & 0.692 \sd{0.003} \\
      \cmidrule(lr){2-6}
      & FR    & PII     & 0.938 \sd{0.030} & 0.923 \sd{0.016} & 0.924 \sd{0.020} \\
      &       & IPI     & 0.526 \sd{0.011} & 0.641 \sd{0.011} & 0.573 \sd{0.010} \\
      &       & Overall & 0.732 \sd{0.016} & 0.782 \sd{0.006} & 0.748 \sd{0.011} \\
    \midrule

    \multirow{12}{*}{mmBERT-base}
      & Mixed & PII     & 0.917 \sd{0.021} & 0.912 \sd{0.008} & 0.914 \sd{0.012} \\
      &       & IPI     & 0.528 \sd{0.006} & 0.565 \sd{0.007} & 0.544 \sd{0.003} \\
      &       & Overall & 0.710 \sd{0.012} & 0.727 \sd{0.006} & 0.717 \sd{0.006} \\
      \cmidrule(lr){2-6}
      & DE    & PII     & 0.926 \sd{0.005} & 0.898 \sd{0.016} & 0.909 \sd{0.012} \\
      &       & IPI     & 0.703 \sd{0.016} & 0.746 \sd{0.017} & 0.718 \sd{0.010} \\
      &       & Overall & 0.833 \sd{0.008} & 0.835 \sd{0.014} & 0.829 \sd{0.009} \\
      \cmidrule(lr){2-6}
      & EN    & PII     & 0.838 \sd{0.010} & 0.877 \sd{0.036} & 0.843 \sd{0.010} \\
      &       & IPI     & 0.517 \sd{0.018} & 0.555 \sd{0.005} & 0.530 \sd{0.009} \\
      &       & Overall & 0.677 \sd{0.010} & 0.716 \sd{0.020} & 0.686 \sd{0.008} \\
      \cmidrule(lr){2-6}
      & FR    & PII     & 0.956 \sd{0.040} & 0.925 \sd{0.016} & 0.933 \sd{0.023} \\
      &       & IPI     & 0.569 \sd{0.037} & 0.650 \sd{0.010} & 0.603 \sd{0.024} \\
      &       & Overall & 0.762 \sd{0.023} & 0.787 \sd{0.009} & 0.768 \sd{0.015} \\
\midrule
    \multirow{12}{*}{GPT-5.4}
      & Mixed & PII     & 0.818 \phantom{\sd{0.000}} & 0.754 \phantom{\sd{0.000}} & 0.766 \phantom{\sd{0.000}} \\
      &       & IPI     & 0.394 \phantom{\sd{0.000}} & 0.344 \phantom{\sd{0.000}} & 0.348 \phantom{\sd{0.000}} \\
      &       & Overall & 0.592 \phantom{\sd{0.000}} & 0.535 \phantom{\sd{0.000}} & 0.543 \phantom{\sd{0.000}} \\
      \cmidrule(lr){2-6}
      & DE    & PII     & 0.804 \phantom{\sd{0.000}} & 0.728 \phantom{\sd{0.000}} & 0.722 \phantom{\sd{0.000}} \\
      &       & IPI     & 0.563 \phantom{\sd{0.000}} & 0.391 \phantom{\sd{0.000}} & 0.440 \phantom{\sd{0.000}} \\
      &       & Overall & 0.703 \phantom{\sd{0.000}} & 0.588 \phantom{\sd{0.000}} & 0.605 \phantom{\sd{0.000}} \\
      \cmidrule(lr){2-6}
      & EN    & PII     & 0.803 \phantom{\sd{0.000}} & 0.814 \phantom{\sd{0.000}} & 0.740 \phantom{\sd{0.000}} \\
      &       & IPI     & 0.412 \phantom{\sd{0.000}} & 0.297 \phantom{\sd{0.000}} & 0.325 \phantom{\sd{0.000}} \\
      &       & Overall & 0.607 \phantom{\sd{0.000}} & 0.556 \phantom{\sd{0.000}} & 0.533 \phantom{\sd{0.000}} \\
      \cmidrule(lr){2-6}
      & FR    & PII     & 0.829 \phantom{\sd{0.000}} & 0.763 \phantom{\sd{0.000}} & 0.788 \phantom{\sd{0.000}} \\
      &       & IPI     & 0.564 \phantom{\sd{0.000}} & 0.461 \phantom{\sd{0.000}} & 0.503 \phantom{\sd{0.000}} \\
      &       & Overall & 0.696 \phantom{\sd{0.000}} & 0.612 \phantom{\sd{0.000}} & 0.645 \phantom{\sd{0.000}} \\
    \bottomrule
  \end{tabular}
  \caption{Macro-averaged partial-match precision (P), recall (R),
  and $F_1$ on the language-balanced 70/15/15 split, including
  \textcolor{sdgray}{sample SD} across five training seeds. GPT-5.4 is evaluated on a single run for each language setting.}
  \label{tab:aggregated-baselines}
\end{table*}

\label[appendix]{app:app_identifiers}

\begin{table*}[th]
\centering
\footnotesize
\setlength{\tabcolsep}{5pt}
\renewcommand{\arraystretch}{1.2}
\begin{tabular}{@{}>{\raggedright\arraybackslash}p{2.7cm}
                  >{\raggedright\arraybackslash}p{4.0cm}
                  >{\raggedright\arraybackslash}p{8.3cm}@{}}
\toprule
\textbf{Label} & \textbf{Example} & \textbf{Notes} \\
\midrule
\multicolumn{3}{@{}l}{\textit{Personally Identifiable Information (PII)}} \\
\addlinespace[2pt]
\textsc{Name} & John Smith, Anna Müller & First/last name, initials, nicknames; names of patients, but also doctors, social contacts, or others. \\
\textsc{Date} & Born on 12.05.1990; admitted 03/14/2022 & Any direct mention of a date. \\
\textsc{Age} & 92 years old & Any explicit age in years for anyone mentioned in a document. \\
\textsc{Geolocation} & 221B Baker Street; 10115 Berlin & Any explicitly named location or address, including zip codes, city/town names, area codes, or federal states. \\
\textsc{Facility} & St.\ Mary's Hospital; Lincoln High School & Explicitly named hospitals/clinics, legal institutional settings, social/residential institutions, and educational institutions. \\
\textsc{Organization} & Worked at Siemens AG & Explicitly named organizations, institutions, or companies that do not fall under \textsc{Facility}. \\
\textsc{Contact\_Phone} & +49 30 12345678 & All kinds of phone numbers, including fax numbers. \\
\textsc{Contact\_Online} & anna.smith@email.com; @anna\_smith & Emails, social media handles, URLs, usernames. \\
\textsc{Code} & PA-2839; DE 279 328 238 & Medical and non-medical identification codes, numbers, or IDs for cases, accounts, documents, devices, etc. \\
\textsc{Biometric} & fingerprint, retina scan & Biometric identifiers such as fingerprints. \\
\textsc{Profession} & She works as a neurosurgeon & Mentions of professions or specific education. Further occupational detail (e.g.\ part-time work, retirement) is annotated under \textsc{Dem.\_Socio}. \\
\midrule
\multicolumn{3}{@{}l}{\textit{Indirect Personal Information (IPI)}} \\
\addlinespace[2pt]
\textsc{Appearance} & Scar across right eye & Distinct physical features. \\
\textsc{Time} & yesterday/tomorrow; ``every Tuesday''; ``twenty years ago'' & Relative, vague, or contextual time references and schedules. \\
\textsc{Age} & very young; late twenties & Indirect descriptions of age. \\
\textsc{Geolocation} & somewhere up north; near the old harbour & Descriptions of locations that are not explicitly named addresses. \\
\textsc{Facility} & hospital, clinic, my GP practice, ward, lab & Generic (not uniquely named) facilities/institutions: healthcare, legal (courts, police stations, prisons, probation offices), social/residential (shelters, group homes), and educational (schools, daycares). \\
\textsc{Disease\_\allowbreak Diagnosis\_\allowbreak Procedure} & Suffers from hemophilia & Rare conditions increase re-identification risk. \\
\textsc{Symptoms} & Sudden temporary paralysis & Unusual symptom combinations. \\
\textsc{Medication} & A rare immunotherapy drug & Uncommon medications. \\
\textsc{Result} & Extremely elevated rare antibody levels & Unexpected or rare results. \\
\textsc{History} & Was narcoleptic as a child & Unique medical life events. \\
\textsc{Legal} & Has a restraining order against person~X & Any mention of legal status, supervision, etc. \\
\textsc{Lifestyle} & Enjoys long-distance running & Habits and routines. \\
\textsc{Context} & Symptoms began after a car accident & Life events in aggregation. \\
\textsc{Demographics} & In a heterosexual relationship but identifies as bisexual & Gender, ethnicity, language, sexual orientation, etc. \\
\textsc{Dem.\_Family\_\allowbreak Household} & Her mother has a history of depression & References to family members without personal names. \\
\textsc{Dem.\_Socio} & student; goes to kindergarten; private insurance & Socioeconomic, employment, or educational status; homelessness; insurance status. \\
\textsc{Quantity} & 20\,mg daily; 12\,km & Meaningful quantities not better captured elsewhere: money amounts, percentages, counts, distances. \\
\bottomrule
\end{tabular}
\caption{Identifier taxonomy used for annotation.
\textsc{Demographics}, \textsc{Demographics\_Family\_Household} and \textsc{Demographics\_Socio} are three separate types that share a prefix, abbreviated as \textsc{Dem.} in all tables.}
\label{tab:identifiers}
\end{table*}

\end{document}